\documentclass[10pt,twocolumn,letterpaper]{article}

\usepackage{cvpr}
\usepackage{times}
\usepackage{epsfig}
\usepackage{graphicx}
\usepackage{amsmath}
\usepackage{amssymb}

\usepackage{paralist}                   
\usepackage{amsmath}                    
\usepackage{algorithm, algpseudocode}   
\usepackage{tabu, multirow}             
\usepackage{color}
\usepackage{soul}
\usepackage{graphicx}
\usepackage{wrapfig}
\usepackage{mathtools}
\usepackage{comment}
\usepackage{caption}
\usepackage{subcaption}
\usepackage{balance}
\usepackage{authblk}

\usepackage[pagebackref=true,breaklinks=true,letterpaper=true,colorlinks,bookmarks=false]{hyperref}

\DeclareMathAlphabet{\pazocal}{OMS}{zplm}{m}{n}
\newcommand{\Dc}{\pazocal{D}}

\newcommand{\Xc}{\pazocal{X}}
\newcommand{\Yc}{\pazocal{Y}}
\newcommand{\Lc}{\pazocal{L}}

\newcommand{\E}{\mathbb{E}}
\newcommand{\norm}[1]{\left\lVert#1\right\rVert}

\newcommand{\ours}{{GADA}}

\DeclarePairedDelimiterX{\infdivx}[2]{(}{)}{%
  #1\;\delimsize\|\;#2%
}
\newcommand{\kl}{\text{D}_\text{KL}\infdivx} 

\newcommand{\myparagraph}[1]{\vspace{-0.25cm} \paragraph{#1}}

\cvprfinalcopy 

\def\cvprPaperID{7615} 

\ifcvprfinal\fi
\begin{document}



\title{Enlarging Discriminative Power by Adding an Extra Class \\ in Unsupervised Domain Adaptation}


\author{Hai H. Tran}
\author{Sumyeong Ahn}
\author{Taeyoung Lee}
\author{Yung Yi}
\affil{School of Electrical Engineering\\
Korea Advanced Institute of Science and Technology}

\newcommand{\real}{{\mathbb R}}
\newcommand{\integer}{{\mathbb Z}}


\newcommand{\td}{\ensuremath{t_{\text{dl}}}}
\newcommand{\bx}{\bm{x}}
\newcommand{\bc}{\bm{c}}
\newcommand{\bbx}{\ensuremath{{\bm x}}}
\newcommand{\bbc}{\ensuremath{{\bm c}}}
\newcommand{\bu}{\ensuremath{{\bm u}}}

\newcommand{\bz}{\ensuremath{{\bm z}}}
\newcommand{\ba}{\ensuremath{{\bm a}}}
\newcommand{\be}{\ensuremath{{\bm e}}}
\newcommand{\br}{\ensuremath{{\bm r}}}
\newcommand{\by}{\ensuremath{{\bm y}}}
\newcommand{\bby}{\ensuremath{{\bm y}}}
\newcommand{\bbq}{\ensuremath{{\bm q}}}

\newcommand{\bp}{\ensuremath{{\bm p}}}
\newcommand{\bh}{\ensuremath{{\bm h}}}
\newcommand{\bP}{\bm{P}}
\newcommand{\bA}{\ensuremath{{\bm A}}}

\newcommand{\bF}{\bm{F}}
\newcommand{\bN}{\ensuremath{{\bm N}}}
\newcommand{\bbf}{\ensuremath{{\bm f}}}
\newcommand{\bG}{\ensuremath{{\bm G}}}
\newcommand{\bQ}{\ensuremath{{\bm Q}}}
\newcommand{\bH}{\ensuremath{{\bm H}}}
\newcommand{\bW}{\ensuremath{{\bm W}}}
\newcommand{\bJ}{\ensuremath{{\bm J}}}
\newcommand{\bg}{\ensuremath{{\bm g}}}
\newcommand{\bs}{\ensuremath{{\bm s}}}

\newcommand{\bbv}{\ensuremath{{\bm v}}}
\newcommand{\bbz}{\ensuremath{{\bm z}}}
\newcommand{\bbu}{\ensuremath{{\bm u}}}
\newcommand{\bbm}{\ensuremath{{\bm m}}}
\newcommand{\bbU}{\ensuremath{{\bm U}}}
\newcommand{\bw}{\bm{w}}
\newcommand{\bbr}{\ensuremath{{\bm r}}}
\newcommand{\bbalpha}{\ensuremath{{\bm \alpha}}}
\newcommand{\brho}{\ensuremath{{\bm \rho}}}



\newcommand{\bR}{\bm{R}}

\newcommand{\cR}{\ensuremath{\mathcal R}}
\newcommand{\set}[1]{\ensuremath{\mathcal #1}}
\renewcommand{\vec}[1]{\bm{#1}}
\newcommand{\mat}[1]{\bm{#1}}

\newcommand{\node}[1]{\ensuremath{{\tt #1}}}
\newcommand{\nn}[1]{\node{#1}}

\newcommand{\mytitle}[1]{\medskip \noindent{\em #1} \smallskip}
\newcommand{\mytitlehead}[1]{\noindent{\em #1} \smallskip}
\newcommand{\bnote}[1]{{\color{blue}#1}} 
\newcommand{\rnote}[1]{{\color{red}#1}} 

\newcommand{\separator}{
  \begin{center}
    \rule{\columnwidth}{0.3mm}
  \end{center}
}
\newenvironment{separation}{ \vspace{-0.3cm}  \separator  \vspace{-0.2cm}}
{  \vspace{-0.4cm}  \separator  \vspace{-0.1cm}}

\def\un{\underline}
\def\ov{\overline}
\def\Bl{\Bigl}
\def\Br{\Bigr}
\def\lf{\left}
\def\ri{\right}
\def\st{\star}

\newcommand{\bprob}[1]{\mathbb{P}\Bl[ #1 \Br]}
\newcommand{\prob}[1]{\mathbb{P}[ #1 ]}
\newcommand{\expect}[1]{\mathbb{E}[ #1 ]}
\newcommand{\bexpect}[1]{\mathbb{E}\Bl[ #1 \Br]}
\newcommand{\bbexpect}[1]{\mathbb{E}\lf[ #1 \ri]}
\newcommand{\vari}[1]{\text{VAR}[ #1 ]}
\newcommand{\bvari}[1]{\text{VAR}\Bl[ #1 \Br]}

\newcommand{\exming}{\ensuremath{\text{ExMin}_g} }
\newcommand{\imp}{\Longrightarrow}
\newcommand{\beq}{\begin{eqnarray*}}
\newcommand{\eeq}{\end{eqnarray*}}
\newcommand{\beqn}{\begin{eqnarray}}
\newcommand{\eeqn}{\end{eqnarray}}
\newcommand{\bemn}{\begin{multiline}}
\newcommand{\eemn}{\end{multiline}}

\newcommand{\sqeq}{\addtolength{\thinmuskip}{-4mu}
\addtolength{\medmuskip}{-4mu}\addtolength{\thickmuskip}{-4mu}}

\newcommand{\unsqeq}{\addtolength{\thinmuskip}{+4mu}
\addtolength{\medmuskip}{+4mu}\addtolength{\thickmuskip}{+4mu}}

\newcommand{\lfl}{{\lfloor}}
\newcommand{\rfl}{{\rfloor}}
\newcommand{\floor}[1]{{\lfloor #1 \rfloor}}

\newcommand{\grad}[1]{\nabla #1}







\def\A{\mathcal A}
\def\oA{\overline{\mathcal A}}
\def\S{\mathcal S}
\def\D{\mathcal D}
\def\eff{{\rm Eff}}

\def\cU{{\cal U}}
\def\cM{{\cal M}}
\def\cV{{\cal V}}
\def\cA{{\cal A}}
\def\cX{{\cal X}}
\def\cN{{\cal N}}
\def\cJ{{\cal J}}
\def\cK{{\cal K}}
\def\cL{{\cal L}}
\def\cI{{\cal I}}
\def\cY{{\cal Y}}
\def\cZ{{\cal Z}}
\def\cC{{\cal C}}
\def\cR{{\cal R}}
\def\id{{\rm Id}}
\def\st{{\rm st}}
\def\cF{{\cal F}}
\def\cG{{\mathcal G}}
\def\N{\mathbb{N}}
\def\R{\mathbb{R}}
\def\cB{{\cal B}}
\def\cP{{\cal P}}
\def\cS{{\cal S}}
\def\ind{{\bf 1}}

\def\bmg{{\bm{\gamma}}}
\def\bmr{{\bm{\rho}}}
\def\bmq{{\bm{q}}}
\def\bmt{{\bm{\tau}}}
\def\bmn{{\bm{n}}}
\def\bmcapn{{\bm{N}}}
\def\bmrho{{\bm{\rho}}}

\def\igam{\underline{\gamma}(\lambda)}
\def\sgam{\overline{\gamma}(\lambda)}

\def\PP{{\mathrm P}}
\def\EE{{\mathrm E}}
\def\iskip{{\vskip -0.4cm}}
\def\siskip{{\vskip -0.2cm}}

\def\bp{\noindent{\it Proof.}\ }
\def\ep{\hfill $\Box$}

\maketitle

\begin{abstract}
In this paper, we study the problem of unsupervised domain adaptation that aims at obtaining a prediction model for the target domain using labeled data from the source domain and unlabeled data from the target domain.
%
There exists an array of recent research based on the idea of extracting features that are not only invariant for both domains but also provide high discriminative power for the target domain.
In this paper, we propose an idea of empowering the discriminativeness:
Adding a new, artificial class and training the model on the data together with the GAN-generated samples of the new class.
The trained model based on the new class samples is capable of extracting the features that are more discriminative by repositioning data of current classes in the target domain and therefore drawing the decision boundaries more effectively. Our idea is highly generic so that it is compatible with many existing methods such as DANN, VADA, and DIRT-T.
We conduct various experiments for the standard data commonly used for the evaluation of unsupervised domain adaptations and demonstrate that our algorithm achieves the SOTA performance for many scenarios.



\end{abstract}

\section{Introduction}
\label{sec:intro}

Deep neural networks have recently been used as a major way of
achieving superb performance on various machine learning tasks, e.g.,
image classification \cite{he2016deep}, image generation \cite{gan14},
and speech recognition \cite{amodei16}, just to name a few.  However,
it still leaves much to be desired when a network trained on a dataset
from a specific data source is used for
dataset from another
data source.  This domain shift and thus distribution mismatch
frequently occurs in practice, and has been studied in the area of
domain adaptation. The crucial ingredient in domain adaptation lies in
transferring the knowledge from the
source domain to the model used in the target domain.


\begin{figure}[!t]
    \captionsetup[subfigure]{justification=centering}
    \hspace{-0.3cm}
    \begin{subfigure}[t]{0.51\columnwidth}
        \centering
        \includegraphics[width=.99\textwidth]{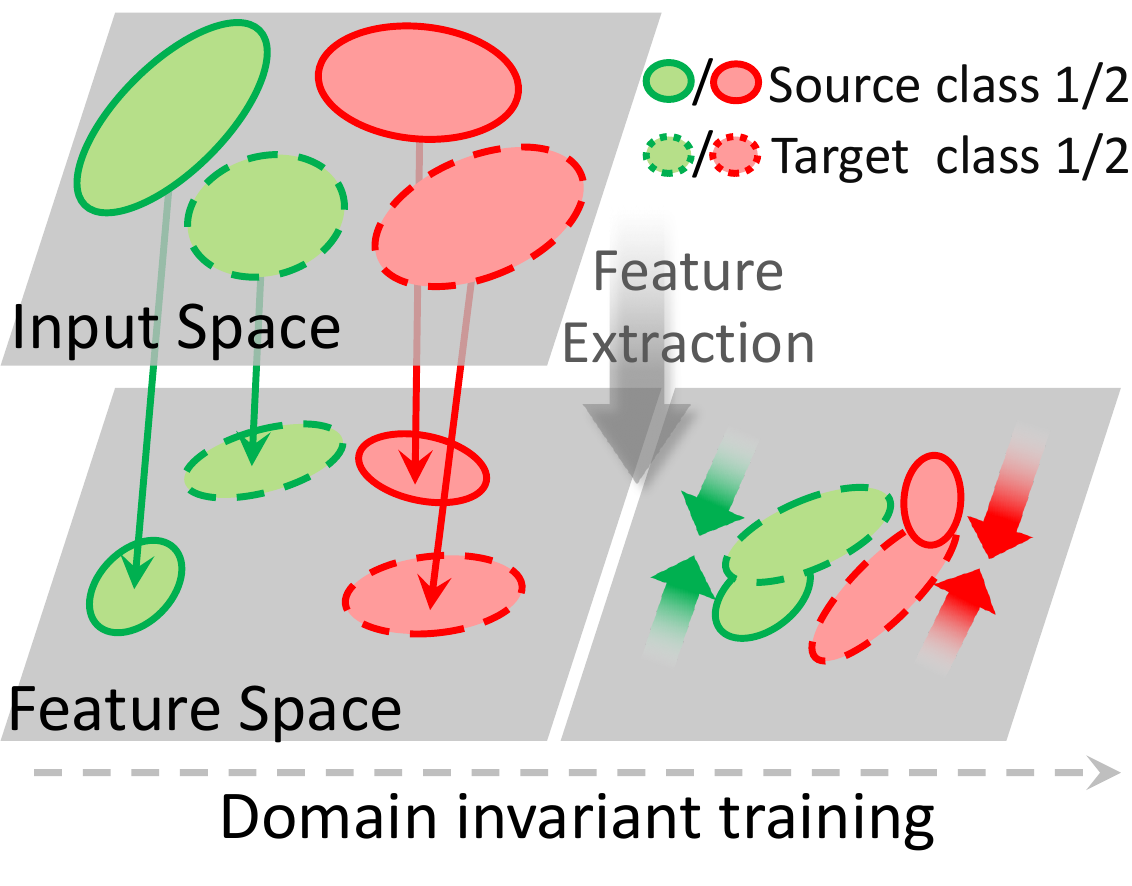}
        \caption{\footnotesize Domain-invariant feature \newline extraction}
        \label{fig:DANN}
    \end{subfigure}
    \hspace{-0.1cm}
    \begin{subfigure}[t]{0.51\columnwidth}
        \centering
        \includegraphics[width=.99\textwidth]{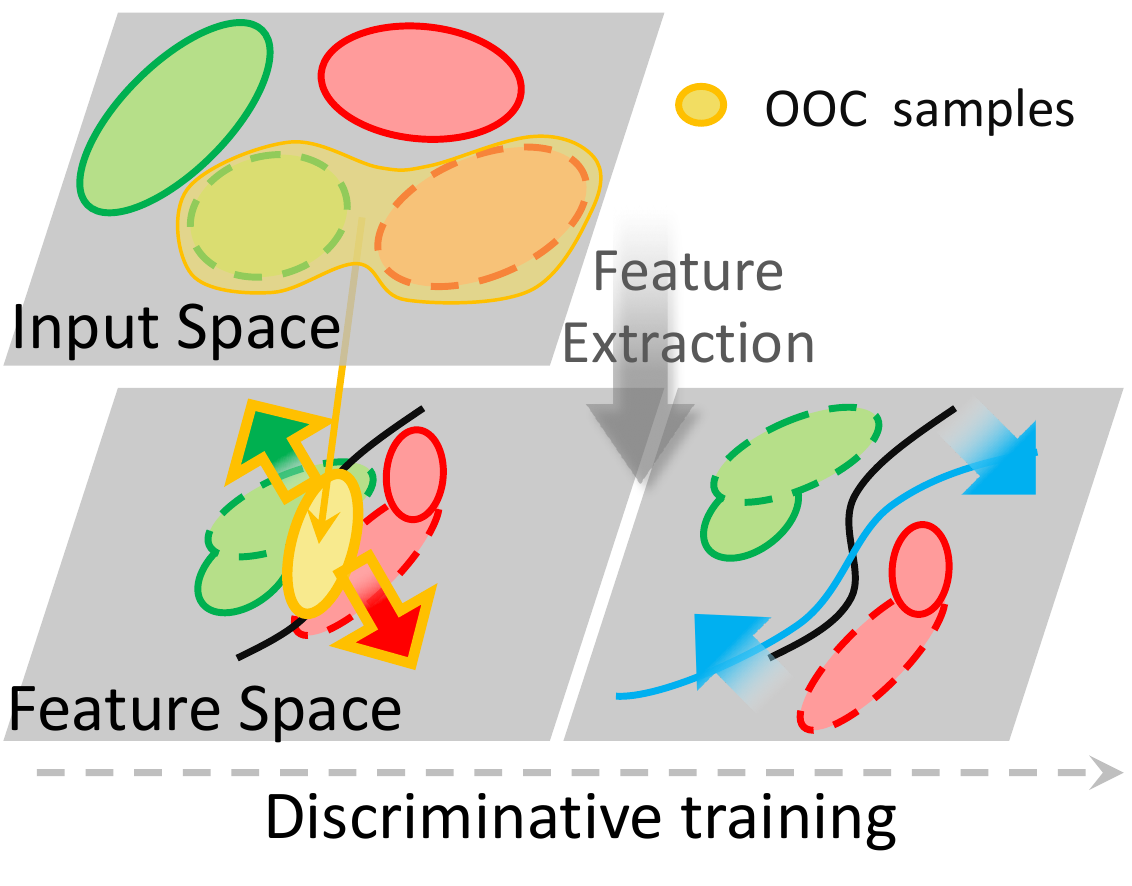}
        \caption{\footnotesize Larger discriminative power: \newline``fictitious'' class and OOC samples.}
        \label{fig:bad_generator}
    \end{subfigure}
    \caption{\small Illustration on how {\ours} works. Each arrow in the feature space corresponds to the force that moves the extracted features or decision boundary. (a) describes how domain-invariant features are learned. (b) explains how discriminative features are extracted by utilizing out-of-class (OOC) samples. The OOC samples and $(K+1)^{\text{th}}$ class increase the distance between ``real'' clusters, which helps the classifier place the decision boundary in the low-density area easier.}
    \label{fig:motivation}
\end{figure}

In this paper, we consider the classification problem of {\em unsupervised} domain
adaptation, where the trained model has no access to any label from the target domain.
What a good domain adapation model has to have is two-fold.
First, it is able to extract domain-invariant features that are
present in both source and target domains, thereby aligning the
feature space distributions between two different domains, e.g.,
\cite{tzeng14, long15, long17, sun16, ganin15, ganin16, dsn16}.
Second, it has to have high discriminative power for the target domain task, which
becomes possible by smartly mixing the
following two operations: (i) extracting task-specific, discriminative features~\cite{adda17,xie18c,kumar18} and
(ii) calibrating the extracted feature space so as to
have a clearer separation among classes, e.g.,
moving the decision boundaries~\cite{shu2018a} (see Section~\ref{sec:related} for more details).

Despite recent advances in unsupervised domain adaptation, there still exists non-negligible performance gap between domain adapted classifiers
and fully-supervised classifiers, hinting a room for further improvement. In this paper, we focus on the second part of empowering the predictive model with more discriminativeness, whose key idea is as follows:
Assuming that there are $K$ classes in the target
data, we equip the model with an extra $(K+1)^{\text{th}}$
class. This extra class is constructed so as
to contain the target samples, which we call out-of-class (OOC) samples throughout this paper,
that fail to belong to any of $K$ classes.
Feeding such  OOC samples
and classifying them into the $(K+1)^{\text{th}}$ class help to
provide the classifier with new samples, thereby improving its feature extraction power in terms of discriminativeness.
Figure~\ref{fig:motivation} illustrates our idea, where
to obtain the OOC samples, we train a generator based on a feature matching GAN \cite{salimans16}. We call our idea GADA (Generative Adversarial Domain Adaptation).

\begin{figure}[!t]
  \centering
  \begin{subfigure}{0.5\columnwidth}
    \centering
    \includegraphics[width=\linewidth]{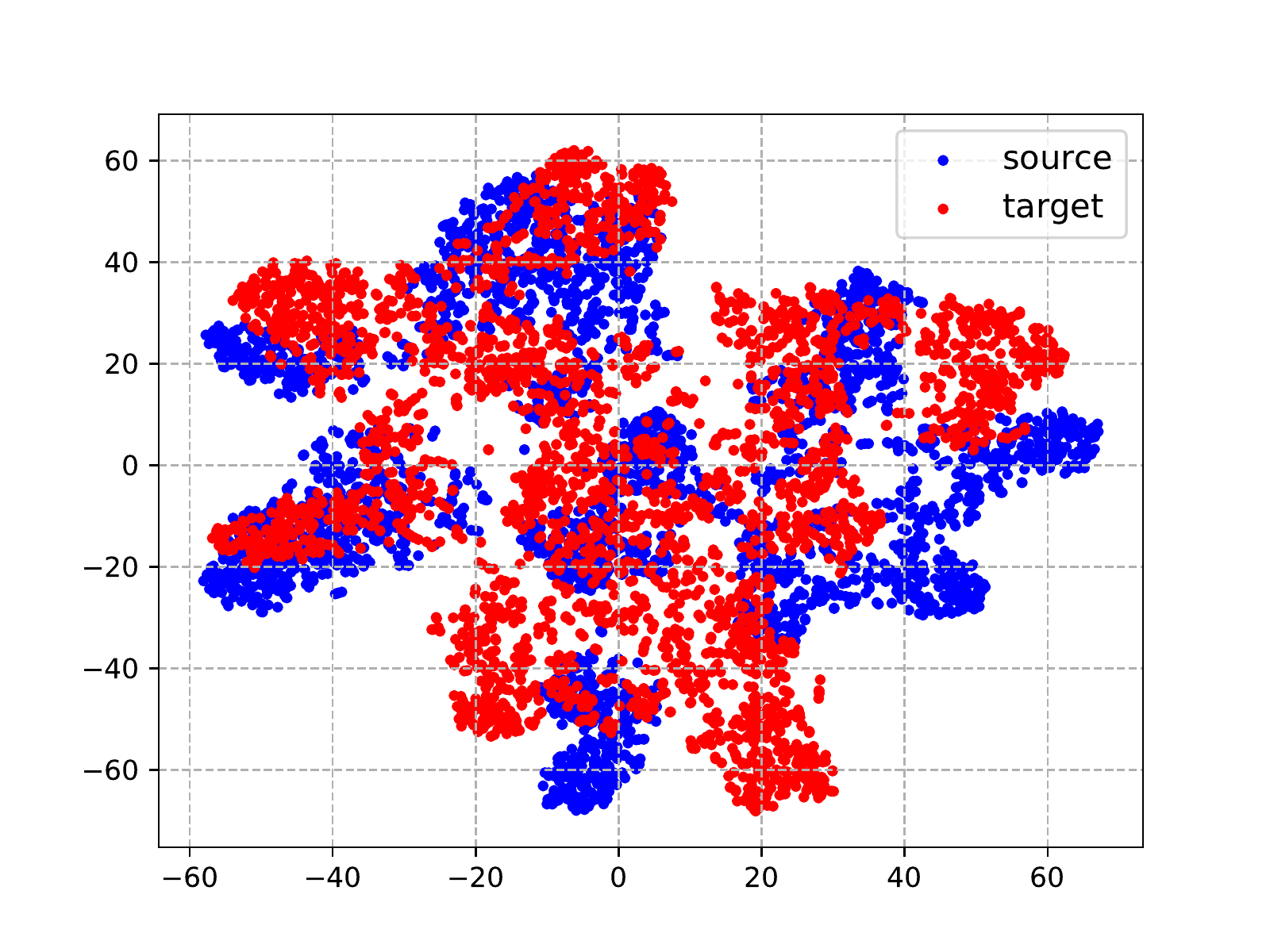}
    \caption{\footnotesize DANN~\cite{ganin16} (74.9\%)}
    \label{fig:feat-dann-stom}
  \end{subfigure}%
  \begin{subfigure}{0.5\columnwidth}
    \centering
    \includegraphics[width=\linewidth]{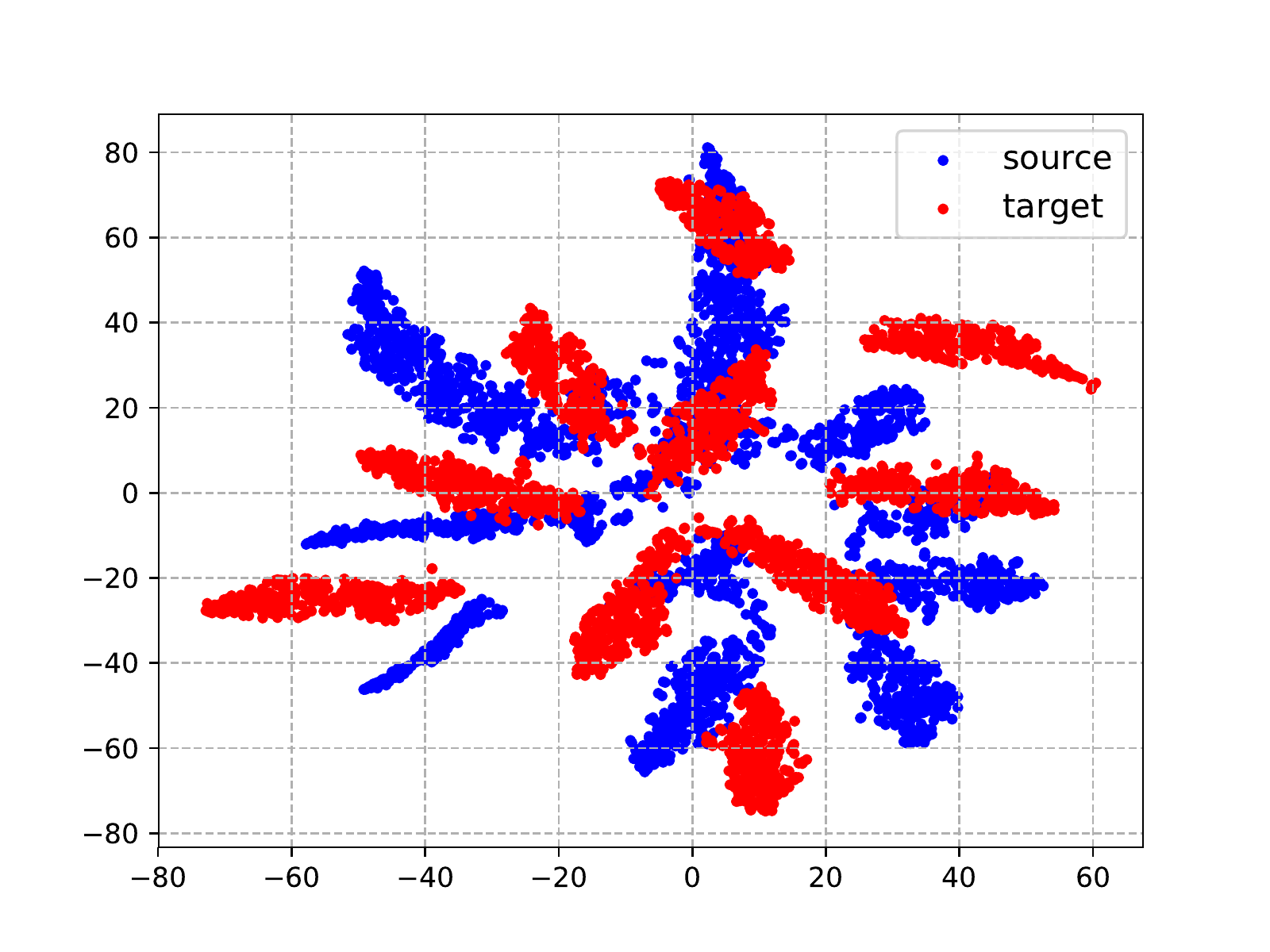}
    \caption{\footnotesize {\ours} (99.0\%)}
    \label{fig:feat-gada-stom}
  \end{subfigure}
  \caption{\small Feature space comparison for the domain adaptation task SVHN $\rightarrow$ MNIST. The number in parenthesis corresponds to the classification accuracy.
  }
  \label{fig:feat-space-intro}
\end{figure}

This power of an extra class has already been verified in the area of semi-supervised learning \cite{salimans16, dai17, qi18}. Our contribution is to apply this idea to  unsupervised domain adaptation in conjunction with necessary engineering components to be practically realized. To the best of our knowledge, this paper is the first to integrate the idea of adding an extra class with unsupervised domain adaptation.
We comment that, compared to the case of semi-supervised learning, it is
necessary to learn both the domain-invariant and the discriminative features, requiring to strike a good balance between those two in domain adaptation.

We highlight that our method is highly generic so as to be compatible with many existing methods.
Figure~\ref{fig:feat-space-intro} shows the feature space illustration, demonstrating the power of {\ours}, when used together with the notorious method, DANN~\cite{ganin16}.
As Figure~\ref{fig:feat-space-intro} shows, we achieve a significant improvement in terms of accuracy and separability among the classes.
We also show our integration power with two recent methods, VADA and DIRT-T~\cite{shu2018a},
which are the methods that improve the model's discriminative power.
VADA aims to extract discriminative features better by employing smart loss functions in training, whereas DIRT-T refines the decision boundary for given extracted features.
As shown later in Section~\ref{sec:exp}, we achieve the best performance
in the most difficult task MNIST $\rightarrow$ SVHN after the integration.
This implies that (i) simply adding a new, fictitious class and training with generated samples as in {\ours} outperforms the VADA algorithm, and (ii) our idea is significantly synergic with a refining-based method DIRT-T.

We empirically prove the effects of our method by carrying out an
extensive set of experiments where we observe that our method outperforms other
state-of-the-art methods on four among six standard domain adaptation tasks,
consisting of the datasets MNIST, SVHN, MNIST-M, DIGITS, CIFAR, and STL.
Although the task SVHN $\rightarrow$ MNIST had a very high accuracy achieved by the existing methods,
{\ours} is demonstrated to surpass all of them.
As for MNIST $\rightarrow$ SVHN, which is known to be extremely challenging,
we integrate our module with VADA~\cite{shu2018a}
to yield an improvement of $13\%$ in terms of accuracy,
thereby setting a new state-of-the-art benchmark.

\section{Related Work}
\label{sec:related}

For presentational convenience, we present the related work by
classifying them into two categories based on their emphasis on (i) extracting
domain-invariant features and (ii) improving discriminativeness.

\myparagraph{Extracing domain-invariant features} A collection of work
\cite{tzeng14, long15, long17, sun16} aimed at aligning the feature
space distributions of the source and target domains by minimizing the
statistical discrepancy between their two distributions using
different metrics.  In~\cite{tzeng14, long15}, maximum mean
discrepancy (MMD) was used to align the high layer feature space.
In \cite{long17}, Joint MMD (JMMD) was used by
defining the distance between the joint distributions of feature space
for each layer one by one. In~\cite{sun16}, the covariances of
feature space were used as the discrepancy to be minimized.
Different approaches include \cite{ganin15} and \cite{artem18}.
The authors in \cite{artem18} proposed a method of minimizing the
regularization loss between the source and target
feature network parameters so as to have similar feature embeddings.
DANN \cite{ganin15} used a domain adversarial neural network, where
the feature extractor is trained to generate domain-invariant features
using a gradient reversal layer,
which inverses the sign of gradients from a domain discriminator.

\myparagraph{Improving discriminativeness}
The idea in DANN has been used as a key component in many subsequent
studies \cite{adda17, xie18c, shu2018a,kumar18}, which essentially modified the adversarial training architecture to acquire more discriminative power.
Different from the end-to-end training in DANN,
ADDA (Adversarial Discriminative Domain Adaptation) \cite{adda17} divided the training into two stages: (a) normal supervised learning on a feature extractor
and a feature classifier on the source domain, and (b) training the target domain's
feature extractor to output the features similar to the source domain's.
In \cite{xie18c}, a semantic loss function is used to measure
the distance between the centroids of the same class from different domains.
Then, minimizing the semantic loss function ensures that the features in the same
class from different domains will be mapped nearby.
VADA (Virtual Adversarial Domain Adaptation)~\cite{shu2018a}
add two loss functions to DANN to move the decision boundaries to low-density regions.
DIRT-T~\cite{shu2018a} solves the non-conservative domain adaptation problem
by applying an additional refinement process to the model trained by VADA.

We summarize other array of work designed for improving discriminativeness.
Tri-training method \cite{saito17a} used high-quality pseudo-labeled samples
to train an exclusive classifier for the target domains via ensemble neural networks.
CoDA (Co-regularized Domain Adaptation)~\cite{kumar18} increases the search space
by introducing multiple feature embeddings using multiple networks, aligning
the target distribution into each space and co-regularizing them to make the networks agree on their predictions.
In GAGL (Generative Adversarial Guided Learning)~\cite{gagl18}, the authors
used a generator trained with CMD (Central Moment Discrepancy)~\cite{cmd17},
similar to what we propose in this paper,
in order to boost the classifier performance.
However, their experiment results are far
from the state-of-the-art performance.

\myparagraph{Pixel-level approach}
We have focused on the feature-level
domain adaptation. There exist pixel-level approaches: In
\cite{bou17}, the authors proposed to adapt the two domains in the
pixel level. The works in \cite{murez18} and \cite{hoffman18a} used
Cycle GAN~\cite{cyclegan} to perform the pixel-level adaptation and
integrate it with the feature-level domain adaptation in the same
model to extract better domain-invariant features.

\myparagraph{Bad GAN}
The idea of using a $(K+1)^{\text{th}}$ output to improve the model performance
was widely used in the semi-supervised learning problem~\cite{salimans16, dai17, qi18}.
The work in \cite{salimans16} was the first that introduces the $(K+1)^{\text{th}}$
output and apply it to the semi-supervised learning problems.
Bad GAN~\cite{dai17} first theoretically and empirically proved
the effectiveness of a bad generator in helping the classifier to learn,
and then designed several loss functions as an attempt to generate bad samples.
While the additional output proved its effects in semi-supervised learning,
we utilize it to solve the problem of unsupervised domain adaptation in this paper.



\section{Method}
\label{sec:algo}

\subsection{Unsupervised Domain Adaptation}
The problem of unsupervised domain adaptation is formulated as follows.
We are given the source dataset with labels
$(\Xc_S,\Yc_S)$ from the source domain $\Dc_S$ and the target dataset
$\Xc_T$ from the target domain $\Dc_T$, but the target data has no labels.
A domain shift between the two domains is assumed, i.e., $\Dc_S \ne\Dc_T$.
The ultimate goal of unsupervised domain adaptation is to learn a good
inference function on the target domain $f: \Xc_T \to \Yc_T$ using the
labeled source data $(\Xc_S,\Yc_S)$ and the unlabeled target data $\Xc_T$.

\subsection{{\ours}}
In this section, we present our method, called {\ours} (Generative Adversarial Domain Adaptation), ranging from the overall network architecture to the detailed algorithm description.

\begin{figure}[!t]
  \centering
  \includegraphics[width=0.48\textwidth]{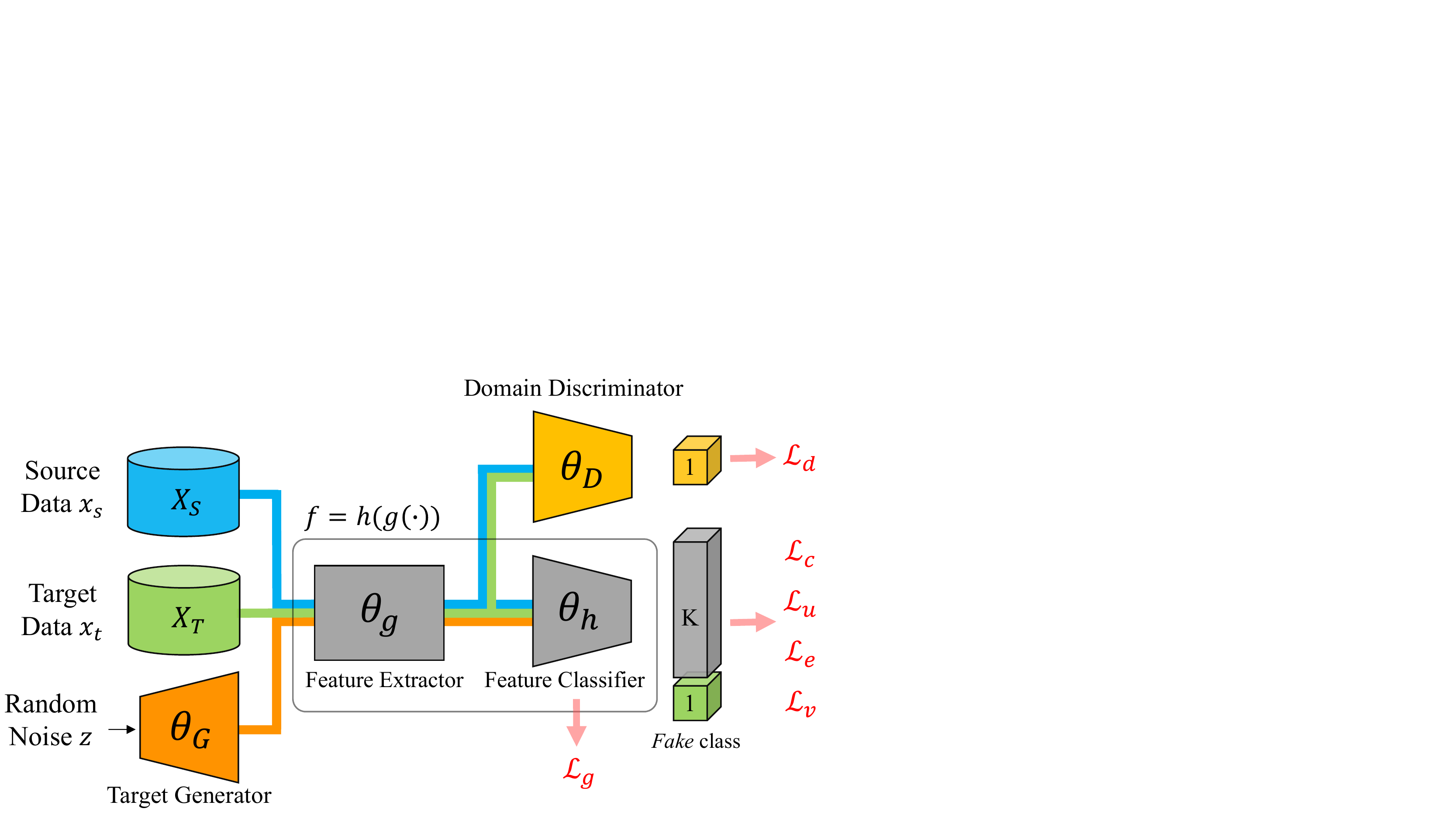}
  \caption{\small Network architecture of GADA. Colored solid lines show the flows of source, target and generated data.
  Six different loss functions are used:
  (i) $\Lc_d$ updates $\theta_D$ and $\theta_g$ for domain-invariance;
  (ii) $\Lc_c$, $\Lc_u$, $\Lc_e$, and $\Lc_v$ updates $\theta_g$ and $\theta_h$ to extract discriminative features, and (iii)
  $\Lc_g$ updates the generator parameters $\theta_G$. The red arrows show the positions where the losses are computed (see Sections~\ref{sec:domain-invariant} and \ref{sec:disc} for details). }
  \label{fig:gada}
\end{figure}

\subsubsection{Network Structure}
We illustrate the network structure of {\ours} in Figure~\ref{fig:gada}, which consists of
four major components {\bf C1-C4} as follows:

\smallskip
\begin{compactenum}[\bf C1.]
\item a feature extractor $g$ with parameters $\theta_g$
\item a feature classifier $h$ with parameters $\theta_h$
\item a domain discriminator $D$ with parameters $\theta_D$
\item a generator $G$ with parameters $\theta_G$
\end{compactenum}
\smallskip


The feature extractor $g$ extracts the common features of the inputs from the source and target domains,
while the feature classifier $h$ classifies the extracted
features from $g$ and outputs the classification scores.
The domain discriminator $D$ is a network with binary output, which indicates
whether an input is from the source domain or the target domain.
The key idea is that, if we are able to fool a smart
discriminator $D$, i.e., making it fail to distinguish the input domains,
the extracted features $g(X)$ become  domain-invariant.
A generator $G$ plays a role of generating the out-of-class (OOC) samples which differs from
the data distribution.
The classifier $f$ is able to distinguish between
the real and the generated OOC samples to have better discriminative power.
This is because when real and OOC samples are separated, the distance between
the clusters of real samples increases, thereby improving
the discriminative quality of the features.
The $(K+1)^{\text{th}}$ class is added to the output layer of the main network $f=h\circ g$,
whose parameter is denoted by $\theta = (\theta_g, \theta_h)$.

\myparagraph{Remark} A couple of remarks are in order.
Firstly, in terms of the network structure, two differences from DANN~\cite{ganin16} exist:
(a) the generator $G$ and (b) the additional $(K+1)^{\text{th}}$ class output.
Second, our method is generic so it can be used with many other approaches
such as DIRT-T \cite{shu2018a}, and CODA \cite{kumar18}, as long as they have their own method of extracting
domain-invariant features.

In the remainder of this section, we elaborate GADA
by separately presenting the parts that contribute to the extraction of domain-invariant and discriminative features in Sections~\ref{sec:domain-invariant} and \ref{sec:disc}, respectively, followed by the whole algorithm description in Section~\ref{sec:whole_algorithm}.

\subsubsection{Domain-invariance via adversarial training}
\label{sec:domain-invariant}
In this subsection, we describe the part of {\ours} which
extracts the features that are invariant for both domains.
This job involves the following three components:
{\bf (C1)} feature extractor $g$, {\bf (C2)} feature classifier $h$ and {\bf (C3)} domain discriminator $D$, where domain-invariant features are extracted by adversarial training.
The key idea is that if we are able to fool a smart
discriminator $D$, i.e., leading $D$ to fail to distinguish the input domains,
the extracted features $g(X)$ turn out to be domain-invariant.

The loss functions\footnote{In this paper, we use the notation $\Lc_x(\theta_y;\Dc_z$) for all loss functions to mean that the loss $\Lc_x$ uses samples from domain $\Dc_z$ to update the parameters $\theta_y$.}
used to train the model are given by:
\begin{align}
  \Lc_c(\theta;\Dc_S) &= \E_{x,y\sim\Dc_S}\left[\log P_\theta(\hat{y}=y|x,y\leq K)\right], \label{eq:lc} \\
  \Lc_d(\theta_g,\theta_D&;\Dc_S,\Dc_T) = \E_{x\sim\Xc_S}\left[\log D(g(x))\right] \cr
  & + \E_{x\sim\Xc_T}\left[\log(1 - D(g(x)))\right],
\end{align}
where $\hat{y}$ indicates the prediction of the network,
$\Lc_d$ is the cross-entropy for the domain discriminator,
and $\Lc_c$ is the negative cross-entropy for the main task\footnote{
We describe $\Lc_c$ as a negative cross-entropy to intuitively show the minimax training mechanism. In the real implementation, $\Lc_c$ is defined as the positive cross-entropy
loss function, so that all the optimization operators are minimization.}.

We note that this is similar to the adversarial training in DANN~\cite{ganin16}, which we also inherit in GADA, as done by other related work \cite{adda17, xie18c, shu2018a,kumar18}. The difference is that
we replace the gradient reversal layer by an
alternating minimization method, which is known to be probably more stable~\cite{shu2018a}.
This alternating training scheme is referred to as
Domain Adversarial Training, and is performed as follows:
\begin{equation}
  \max_{\theta}\min_{\theta_D} \Big [ \Lc_c(\theta;\Dc_S) + \lambda_d \Lc_d(\theta_g,\theta_D;\Xc_S,\Xc_T) \Big ],
\end{equation}
where $\lambda_d$ is the weight of domain discriminator loss $\Lc_d$.
However, the domain discriminator does not consider the class labels while being trained,
so the extracted features are not ensured to have sufficient classification capability.
Therefore, more optimizations are necessary to extract discriminative features,
thereby boosting the performance, which is the key contribution of this paper,
as presented in the next section.

\subsubsection{Discriminativeness by adding a new class}
\label{sec:disc}
We now present how we improve the power of discriminativeness in GADA.
The three components are associated with this process: {\bf (C1)} feature extractor $g$, {\bf (C2)} feature classifier $h$, and {\bf (C4)} generator $G$ (see Figure~\ref{fig:gada}).

\myparagraph{Adding a fictitious class and out-of-class sample generator}
As presented  previously, an OOC (Out-Of-Class) generator generates the samples whose
distribution differs from the target data distribution, which provides the power of
extracting discriminative features from both domains.
In addition, the classifier $f$ must be able to distinguish between
the real and generated samples to have better performance, where
when real and OOC samples are separated, the distance between
the clusters of real samples are increased, thereby improving
the discriminative quality of the features.


In order to help the classifier to distinguish the real and OOC samples,
we introduce an \textit{unsupervised} objective function as follows:
\begin{align}
  \Lc_u(\theta;\Xc_T,&P_z) = \E_{x\sim\Xc_T}\left[\log P_\theta(\hat{y}\leq K|x)\right] \cr
                     & + \E_{z\sim P_z}\left[\log P_\theta(\hat{y}=K+1|G(z))\right],
\end{align}
where $P_z$ is a random noise distribution from which the noise vector $z$ comes.
The function $\Lc_u$ has two terms: (i) the first term is used to
train the network with the unlabeled target data, and
(ii) the second term is to train the network with the generated samples.
By maximizing the first term, we maximize the probability that
an unlabeled target sample belongs to one of the first $K$ classes.
By maximizing the second term, we maximize the probability that
a generated sample belongs to the fictitious $(K+1)^{\text{th}}$ class.

In addition to the objective function used in training the discriminator,
we need a loss function to train a OOC generator.
In~\cite{dai17}, a complementary generator is proposed as a ``perfectly bad" generator which generates no in-distribution samples. However, it is too costly to implement it.
In our model, we use an imperfect complementary generator to reduce the
implementation complexity named Feature Matching (FM) generator~\cite{salimans16}.
The FM generator is trained by minimizing the
feature matching loss function defined as follows:
\begin{multline}
  \Lc_g(\theta_G;\Xc_T,P_z) = \cr \norm{\E_{x\sim\Xc_T}\left[\phi(x)\right] -
  \E_{z\sim P_z}\left[\phi(G(z))\right]},
  \label{eq:fm-loss}
\end{multline}
where $\phi$ is an immediate layer in the network.
In our implementation, we choose $\phi$ to
be the last hidden layer of the feature classifier $h$.
FM matches the statistics (in this case, the mean)
of each minibatch, which leads to a less
constrained loss function that helps the generator
to generate OOC samples~\cite{salimans16,dai17}.
Note that we apply \eqref{eq:fm-loss} to generate the target domain samples only,
because the source samples are provided with the labels,
which are more adequate for training. In addition,
training the network with the generated source samples might
hurt the performance because of non-conservativeness of
domain adaptation \cite{shu2018a} considered in this paper.

\myparagraph{Entropy minimization and virtual adversarial training (VAT)}
We also minimize the entropy of the model's output in order to make the model more
confident about its prediction using the following objective:
\begin{equation}
  \Lc_e(\theta;\Dc_T) = -\E_{x\sim\Dc_T}\left[f(x)^\top \ln f(x)\right].
\end{equation}
This loss prevents the
target data from being located near the decision boundary.
Therefore, it helps the classifier to learn more discriminative
features by placing the samples of the same class closer
to each other in the feature space.

Adversarial training has been proposed to increase the robustness of the classifier
to the adversarial attack which intentionally perturbes samples to degrade
the prediction accuracy. Virtual Adversarial Training (VAT) was proposed for
the same purpose: it ensures consistent predictions for all samples
that are slightly perturbed from the original sample, where the following loss function is used:
\begin{equation}
  \Lc_v(\theta;\Dc) = \E_{x\sim\Dc}\left[\max_{\norm{r}\leq\epsilon}\kl{f(x)}{f(x+r)}\right].
\end{equation}
This loss regularizes
the classifier so that it does not change its prediction abruptly
due to the perturbation of inputs, which helps to learn a robust
classifier. Note that entropy minimization and VAT are popularly used in domain adaptation,
as in~\cite{shu2018a, kumar18}.

\myparagraph{Aggregation}
To extract the discriminative features, using the loss functions introduced earlier, we  perform alternating optimization between the following two:
\begin{align*}
    \max_\theta \ & \Lc_c(\theta;\Dc_S) + \lambda_u\Lc_u(\theta; \Xc_T, P_z) + \lambda_s\Lc_v(\theta;\Dc_S) \cr
    & \hspace{2cm}+ \lambda_t\left[\Lc_v(\theta;\Dc_T) + \Lc_e(\theta;\Dc_T)\right], \cr
  \min_{\theta_G} \ &  \Lc_g(\theta_G; \Xc_T, P_z),
\end{align*}
where $\Lc_c$ is the negative cross-entropy function defined in~\eqref{eq:lc}, while
$\lambda_u$, $\lambda_s$, and $\lambda_t$ are the hyperparameters to control
the impact of each loss function.
Note that the VAT objective function is applied to both the source and target domains, as suggested by \cite{shu2018a}.

\begin{algorithm}[!t]
  \caption{\bf {\ours}}
  \label{alg}
The following three steps are sequentially repeated until convergence.

\smallskip
\begin{compactenum}[\bf S1.]
\item {\bf \em Update the classifier.} Sample $M$ source samples with the corresponding labels
      $(x_S,y_S)$, $M$ unlabeled target samples $x_T$, and $M$ random
      noise vectors $z$, to update the feature extractor $g$ and the feature classifier $h$:
        \begin{align*}
            \max_{\theta}\Lc_c(\theta; \Dc_S) &+ \lambda_d\Lc_d(\theta_g, \theta_D; \Xc_S, \Xc_T) \\
            &+ \lambda_u\Lc_u(\theta;\Xc_T,P_z) + \lambda_s\Lc_v(\theta;\Dc_S) \\
            &+ \lambda_t\left[\Lc_v(\theta;\Dc_T) + \Lc_e(\theta;\Dc_T)\right].
        \end{align*}
\item {\bf \em Update the domain discriminator.} Sample $M$ source samples $x_S$ and $M$ target samples
      $x_T$ to update the domain discriminator $D$ by minimizing $\Lc_d$:
$$\min_{\theta_D}\Lc_d(\theta_g,\theta_D;\Xc_S, \Xc_T).$$
\item {\bf \em Update the generator.} Sample $M$ random noise vectors $z$ and $M$ target
      samples $x_T$, update the generator $G$ by minimizing $\Lc_g$:
      $$\min_{\theta_G} \Lc_g(\theta_G;\Xc_T,P_z).$$
\end{compactenum}





\end{algorithm}

\subsubsection{GADA: Algorithm description}
\label{sec:whole_algorithm}
Combining the two parts in the previous two subsections,
GADA aims at solving the following optimization in training based on the network
structure in Figure~\ref{fig:gada}:
\begin{multline}\label{eq:main-obj}
    \max_{\theta} \min_{\theta_D} \min_{\theta_G}
    \underbrace{\Lc_c(\theta; \Dc_S)}_\text{(a)} +
    \underbrace{\lambda_d\Lc_d(\theta_g,\theta_D;\Xc_S, \Xc_T)}_\text{(b)} \cr
     + \underbrace{\lambda_s\Lc_v(\theta;\Dc_S) + \lambda_t\left[\Lc_v(\theta;\Dc_T) + \Lc_e(\theta;\Dc_T)\right]}_\text{(c)} \cr
     + \underbrace{\lambda_u\Lc_u(\theta;\Xc_T,P_z) + \Lc_g(\theta_G;\Xc_T,P_z)}_\text{(c)}.
\end{multline}
The above function is interpreted as follows.
Maximizing (a) guides the network to achieve
the classification power from the source data and labels.
Updating $\theta_D$ to minimize (b), while
updating $\theta_g$ to maximize it, helps the network
to extract domain-invariant features, as explained in Section~\ref{sec:domain-invariant}.
(c) improves discriminativeness by
generating OOC samples and classifying them into
the fictitious class $K+1$, as well as regularizing the model
with entropy minimization and VAT objective.
The complete training algorithm is presented in Algorithm~\ref{alg}.
Since the algorithm monotonically decreases the objective
function value, the convergence is guaranteed.


\section{Experimental Results}
\label{sec:exp}

\subsection{Domain Adaptation Tasks}
We evaluate our method for the standard
datasets, which include digit datasets
(MNIST, SVHN, MNIST-M, and SynthDigits) and
object datasets (CIFAR-10 and STL-10).

\myparagraph{MNIST $\leftrightarrow$ SVHN}
Both MNIST and SVHN are digit data sets, which differ in style.
MNIST consists of gray-scale hand-written images, while SVHN includes images of RGB house numbers.
Due to the lower image dimension in MNIST,
we upscale MNIST images so as to have the same dimension as SVHN ($32\times 32$) with three same color channels.
The task MNIST $\rightarrow$ SVHN is known to be highly challenging one among the digit adaptation experiments, where
we observe that this task has been omitted in many related papers, possibly due to the adaptation hardness.
The task of the opposite direction SVHN $\rightarrow$ MNIST is relatively easy, compared to MNIST $\rightarrow$ SVHN,
because the test domain MNIST is easier to classify, and
the classifier is trained with the labels from the more complex data set SVHN.

\myparagraph{MNIST $\rightarrow$ MNIST-M}
MNIST-M is constructed by blending the gray-scale MNIST images with colored backgrounds in BSDS500 dataset~\cite{bsds500}. The resulting color images in MNIST-M increase the domain shift between the two datasets, thus
this adaptation task has been widely used to compare the performance of various
models~\cite{ganin16,dsn16,saito17a,shu2018a,kumar18}.

\myparagraph{SynthDigits (DIGITS) $\rightarrow$ SVHN}
SynthDigits is a synthetic digit dataset consisting of 500,000 images generated from Windows fonts
by varying the text, positioning, orientation, background, stroke color, and the amount of blur.
This task reflects a common adaptation task from synthetic images (synthesized images) to real images (house number pictures).

\myparagraph{CIFAR-10 $\leftrightarrow$ STL-10}
Both CIFAR-10 and STL-10 are RGB images, each with 10 different classes.
We remove the non-overlapping class in each data set
(\textit{frog} in CIFAR-10 and \textit{monkey} in STL-10)
and perform the training and evaluation on the 9 leftover classes.
STL-10 has $96\times 96$ image dimension, so we downscale all STL images
to match the $32\times 32$ dimension of CIFAR-10.
Since CIFAR-10 has more labeled data than STL-10, it is easier to adapt from CIFAR-10 to STL-10 than the opposite direction.

\begin{table*}[!t]
    \centering
    \caption{Comparison of state-of-the-art methods for classification accuracy (\%).
    Values in bold are the best.}
    \label{table:main-result}
    \begin{tabu}{ccccccc}
        \tabucline[1pt]{-}
        \multicolumn{1}{r|}{Source}                       & MNIST      & SVHN       & MNIST      & DIGITS     & CIFAR      & STL        \\
        \multicolumn{1}{r|}{Target}                       & SVHN       & MNIST      & MNIST-M    & SVHN       & STL        & CIFAR      \\ \tabucline[1pt]{-}
        \multicolumn{1}{c|}{DANN}          & 35.7       & 71.1       & 81.5       & 90.3       & -          & -          \\
        \multicolumn{1}{c|}{DSN}             & -          & 82.7       & 83.2       & 91.2       & -          & -          \\
        \multicolumn{1}{c|}{ATT}          & 52.8       & 86.2       & 94.2       & 92.9       & -          & -          \\
        \multicolumn{1}{c|}{MSTN}           & -          & 91.7       & -          & -          & -          & -          \\
        \multicolumn{1}{c|}{MCD}           & -          & 96.2       & -          & -          & -          & -          \\
        \multicolumn{1}{c|}{VADA}         & 73.3       & 97.9       & 97.7       & 94.9       & 80.0       & 73.5       \\
        \multicolumn{1}{c|}{VADA+DIRT-T}  & 76.5       & 99.4       & 98.9       & 96.2       & -          & 75.3       \\
        \multicolumn{1}{c|}{CoDA}          & 81.7       & 98.8       & 99         & 96.1       & {\bf 81.4} & 76.4       \\
        \multicolumn{1}{c|}{CoDA+DIRT-T}   & 88.0       & 99.4       & 99.1       & 96.5       & -          & {\bf 77.6} \\
        \multicolumn{1}{c|}{Ours}                         & 83.6       & 99         & 98.8       & 95.9       & 79.7       & 75.1       \\
        \multicolumn{1}{c|}{Ours+DIRT-T}                  & {\bf 90.0} & {\bf 99.6} & {\bf 99.2} & {\bf 96.7} & -          & 76.5       \\ \tabucline[1pt]{-}
    \end{tabu}
\end{table*}

\begin{figure*}[!t]
  \centering
  \begin{subfigure}{0.25\textwidth}
    \centering
    \includegraphics[width=0.85\linewidth]{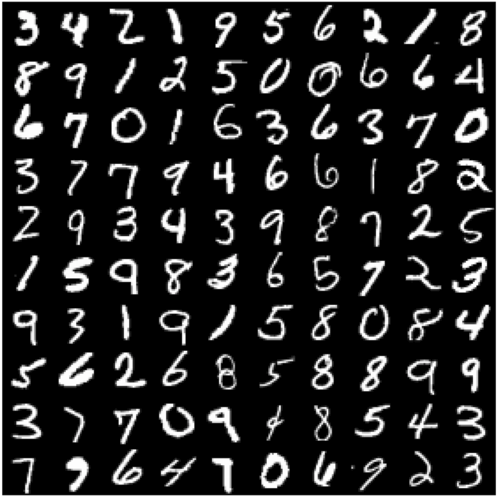}
    \caption{Original MNIST images}
    \label{fig:gen-a}
  \end{subfigure}%
  \begin{subfigure}{0.25\textwidth}
    \centering
    \includegraphics[width=0.85\linewidth]{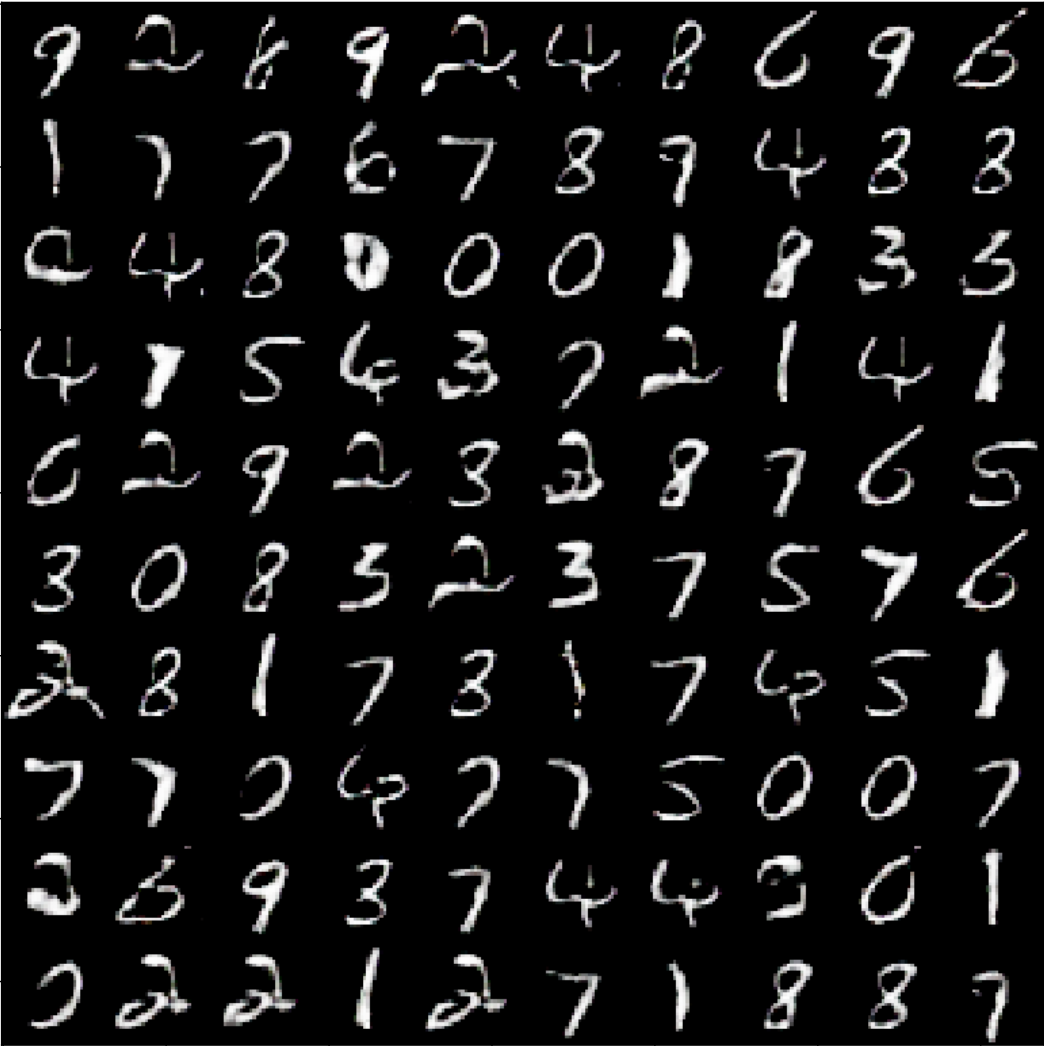}
    \caption{Generated MNIST images}
    \label{fig:gen-b}
  \end{subfigure}%
  \begin{subfigure}{0.25\textwidth}
    \centering
    \includegraphics[width=0.85\linewidth]{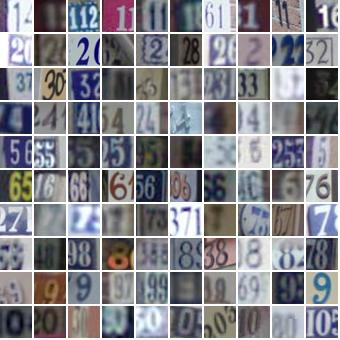}
    \caption{Original SVHN images}
    \label{fig:gen-c}
  \end{subfigure}%
  \begin{subfigure}{0.25\textwidth}
    \centering
    \includegraphics[width=0.85\linewidth]{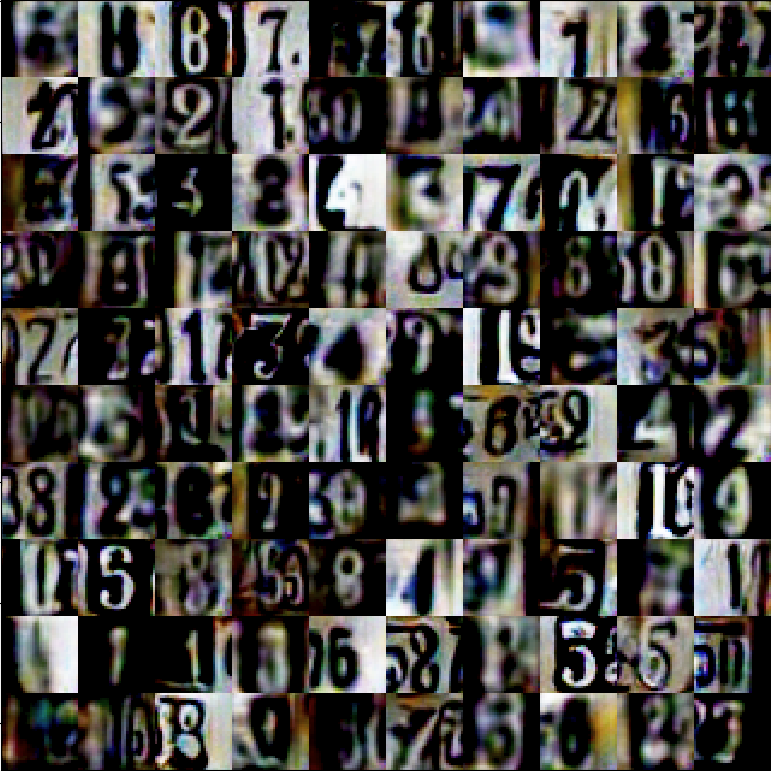}
    \caption{Generated SVHN images}
    \label{fig:gen-d}
  \end{subfigure}
  \caption{Comparison between original and generated images in the tasks
  SVHN $\rightarrow$ MNIST (Figures~\subref{fig:gen-a} and~\subref{fig:gen-b})
  and MNIST $\rightarrow$ SVHN (Figures~\subref{fig:gen-c} and~\subref{fig:gen-d}).
  Bad samples of images are generated after training.}
  \label{fig:gen-images}
\end{figure*}

\subsection{Implementation and Tested Model}


\paragraph{Tested models}
In order to evaluate our method GADA, we compare it against other state-of-the-art algorithms.
They include DANN~\cite{ganin15,ganin16}, DSN~\cite{dsn16},
ATT~\cite{saito17a}, MSTN~\cite{xie18c}, and MCD~\cite{saito18}.
We also contain two recent state-of-the-art methods,
VADA+DIRT-T~\cite{shu2018a} and CoDA~\cite{kumar18}.


We now summarize how to implement our model GADA.
For reproducibility, the source code is given\footnote{\url{https://github.com/haitran14/gada}}.
We refer the readers to the supplementary material for more implementation details.

\plparsep=0.25in
\plitemsep=-0.2in

\begin{compactitem}[$\circ$]
\item {\bf \em Network architecture.}
We use a small convolutional neural network (CNN) for the digit
datasets, and a larger one for the object datasets.
We apply batch normalization to all fully-connected and CNN layers,
while dropout and additive Gaussian noise are used in several layers.
As for the generator, we use transposed convolution
layers to upsample the feature maps.


\item {\bf \em Hyperparameters.}
In all the experiments, we train the network using Adam Optimizer.
We do our hyperparameter search with the learning rate restricted to
$\{2\times10^{-4},10^{-3}\}$, while $\lambda_d$ is either $10^{-2}$ or $0$.
We also restrict other hyperparameters to $\lambda_s=\{0,1\}$,
$\lambda_t=\{10^{-1},10^{-2}\}$ and $\lambda_u=\{10^{-1},10^{-2}\}$.

\item{\bf \em Instance normalization.}
As suggested in \cite{shu2018a}, we apply the instance
normalization to the rescaled input images.
This procedure renders the classifier invariant to channel-wide shifts
and rescaling of pixel intensities.
We choose to apply the normalization process to the tasks
MNIST $\leftrightarrow$ SVHN, and DIGITS $\rightarrow$ SVHN.
We observe that instance normalization is especially crucial
for the task MNIST $\rightarrow$ SVHN, as the classifier performs
extremely bad without the normalization.

\item{\bf \em DIRT-T integration.}
For fair comparison with VADA and CoDA,
after training a model using {\ours},
we refine it using the idea of DIRT-T, which proves to be effective
in improving the performance.
In all the experiments, we refine the model with $\beta=10^{-2}$,
except for STL-10 $\rightarrow$ CIFAR-10, where $\beta$ is set to $10^{-1}$.
Note that we do not apply DIRT-T to CIFAR-10 $\rightarrow$ STL-10
because the number of target samples in the task is low
($450$ samples of STL-10 images), which provides unreliable
estimation of the entropy for minimization.

\item{\bf \em Generator pretraining.}
For the adaptation tasks CIFAR-10 $\leftrightarrow$ STL-10, we pretrain the
feature matching generator before using it to train the classifier
as the noisy gradients at the beginning of the training process would
hurt the training of the generator, especially in these more complicated datasets.
When we start training the main classifier with the pretrained generator,
we keep finetuning the generator with a small learning rate, at $2\times 10^{-5}$.

\end{compactitem}

\begin{figure*}[!t]
  \centering
  \begin{subfigure}{0.26\textwidth}
    \centering
    \includegraphics[width=\linewidth]{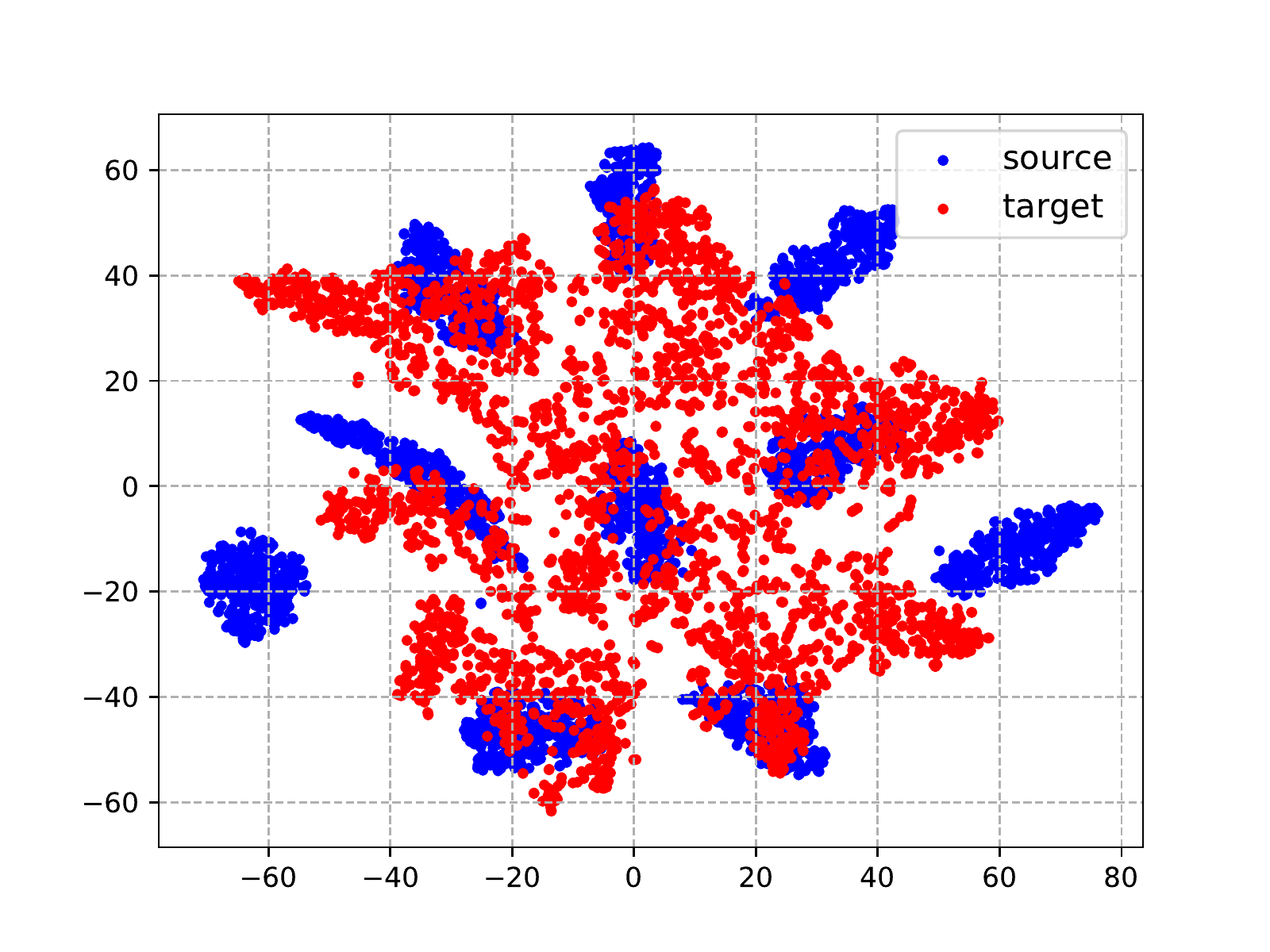}
    \caption{VADA (acc: 70.6\%)}
    \label{fig:feat-vada}
  \end{subfigure}%
  \begin{subfigure}{0.26\textwidth}
    \centering
    \includegraphics[width=\linewidth]{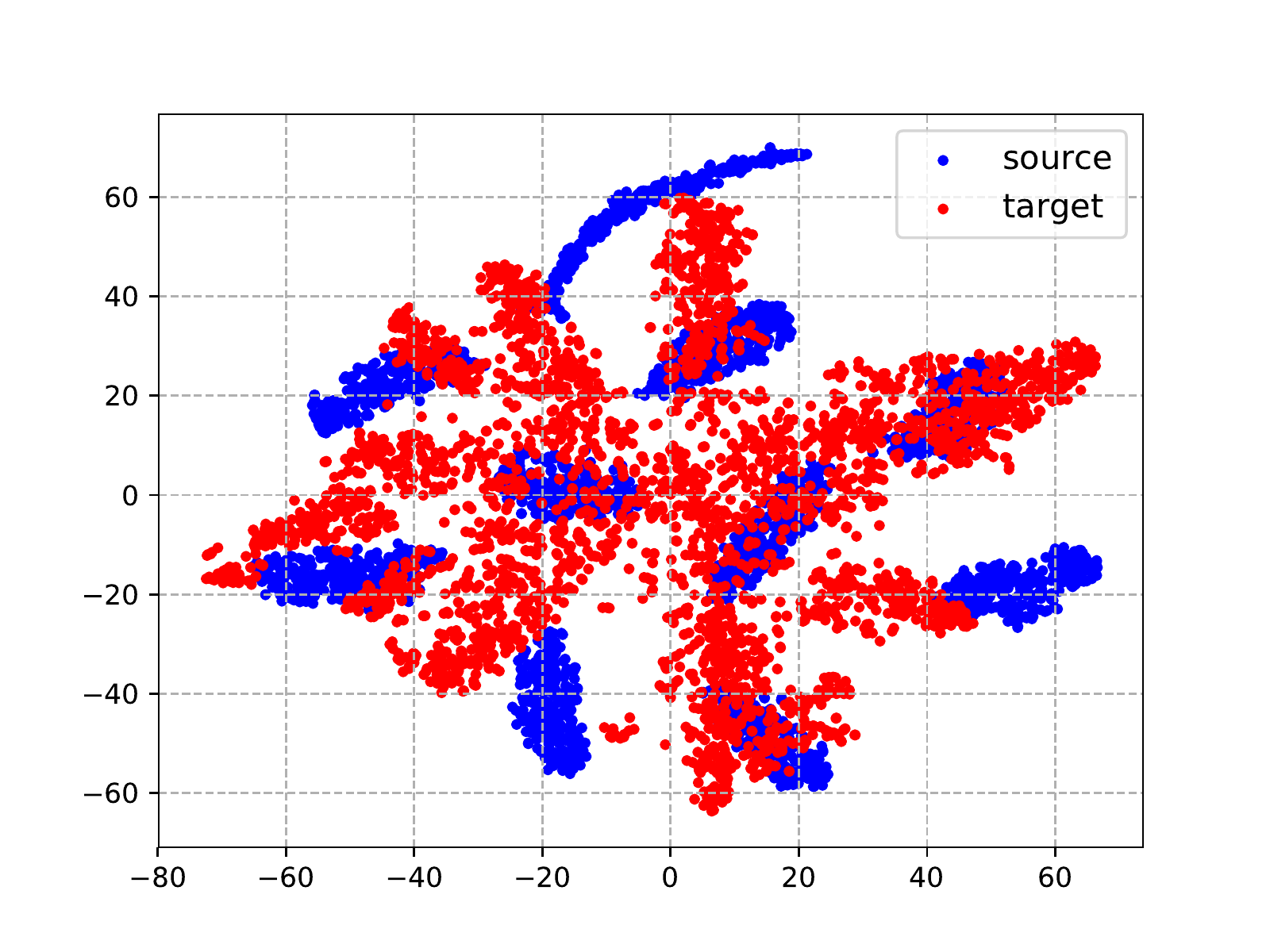}
    \caption{{\ours} (acc: 83.6\%)}
    \label{fig:feat-gada}
  \end{subfigure}%
  \begin{subfigure}{0.26\textwidth}
    \centering
    \includegraphics[width=\linewidth]{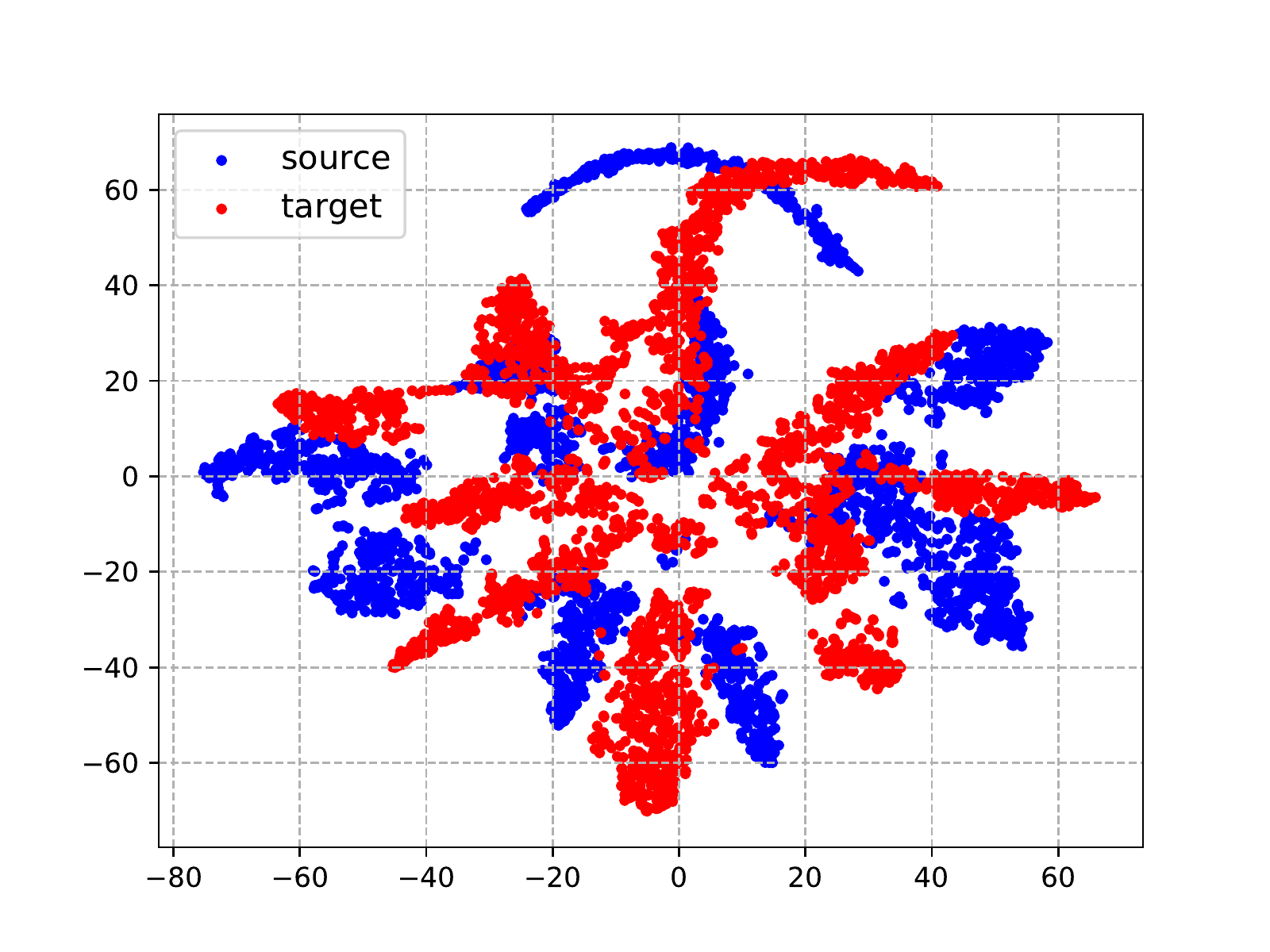}
    \caption{VADA+DIRT-T (acc: 75.75\%)}
    \label{fig:feat-vada-dirtt}
  \end{subfigure}%
  \begin{subfigure}{0.26\textwidth}
    \centering
    \includegraphics[width=\linewidth]{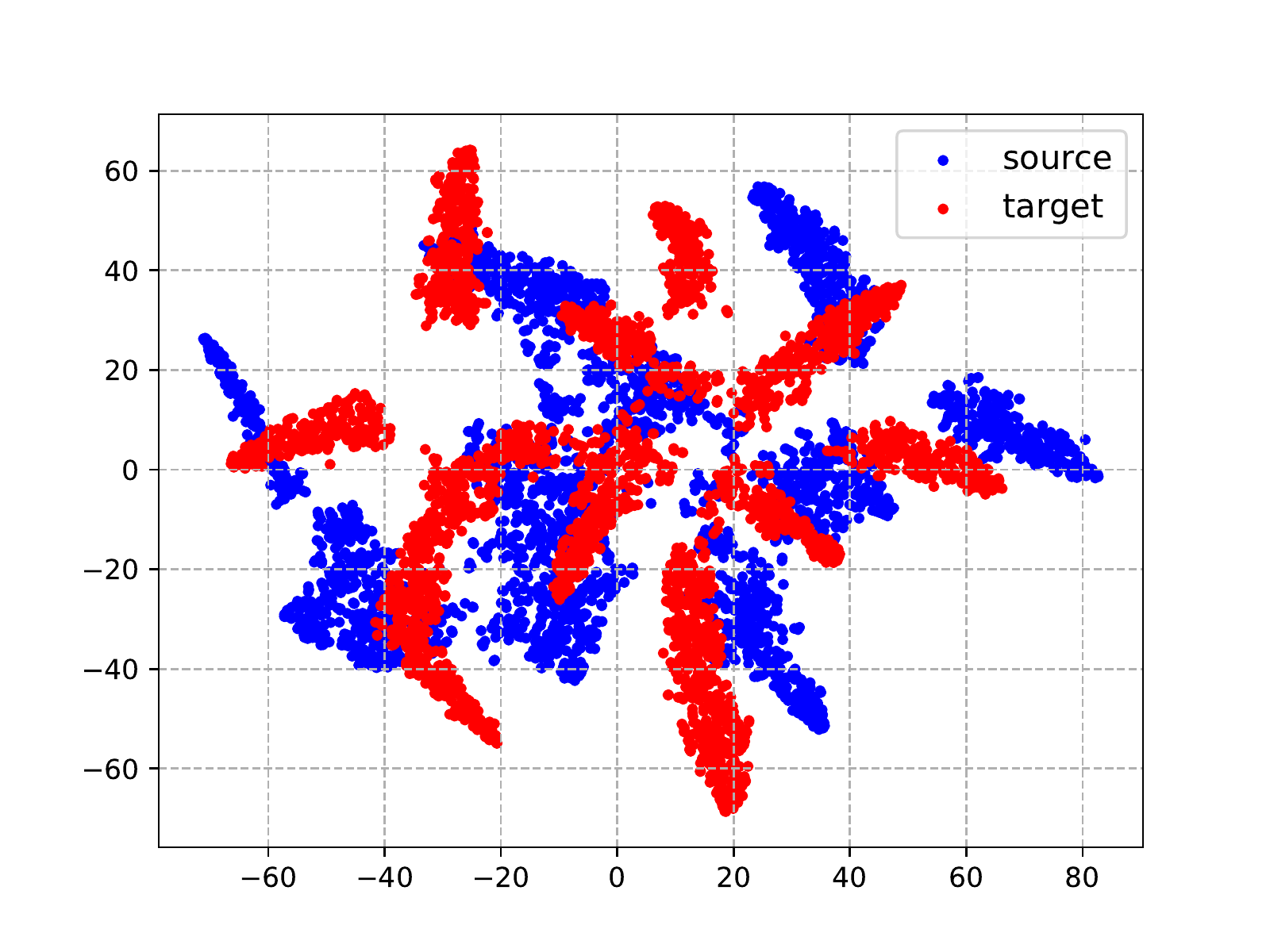}
    \caption{{\ours}+DIRT-T (acc: 90\%)}
    \label{fig:feat-gada-dirtt}
  \end{subfigure}
  \caption{Feature space comparison between VADA~\cite{shu2018a} and {\ours}.
  Combining DIRT-T with {\ours} significantly improves the performance.
  This proves that our {\ours} module could be used to boost other techniques.}
  \label{fig:feat-space}
\end{figure*}

\subsection{Evaluation and Analysis}

\paragraph{Overall comparison}
All the results on comparison with other tested models are presented in Table~\ref{table:main-result}.
To summarize, we achieve state-of-the-art results across four tasks,
SVHN $\rightarrow$ MNIST, MNIST $\rightarrow$ SVHN,
MNIST $\rightarrow$ MNIST-M, and DIGITS $\rightarrow$ SVHN.
Prior methods have achieved very high
performance on the task SVHN $\rightarrow$ MNIST, but GADA outperforms their algorithms.
Note that we only utilize $\Lc_u$ for this configuration.
For the highly challenging adaptation task of MNIST $\rightarrow$ SVHN,
we achieved considerable improvement of approximately $2\%$
over the state-of-the-art algorithm CoDA~\cite{kumar18}.
GADA fails to outperform the SOTA result in CIFAR $\rightarrow$ STL
by a small margin, because STL contains a very small number
of samples in the training set (50 images per class), which seems to hurt
the generator training process.
For STL $\rightarrow$ CIFAR, our performance underpeforms the SOTA
by merely about $1\%$ because the number of labels given for STL training images
is too small, being insufficient for the training.
Overall, we set new state-of-the-art benchmarks in four of the six configurations.

\myparagraph{Generated images}
Generated images are shown in Figure~\ref{fig:gen-images}
for MNIST $\rightarrow$ SVHN and SVHN $\rightarrow$ MNIST.
We see that in both tasks the numbers in the generated images are
recognizable, but the shapes, styles or colors were changed.
This causes them to look different from the original training images, or simply ``bad''.
This analysis empirically proves that the distribution of the generated images is different from the training data's, while keeping meaningful features for the network to learn from.

\myparagraph{Feature space visualization}
In Figure~\ref{fig:feat-space}, we compare the T-SNE plots of the last hidden layer of VADA models (Figures~\ref{fig:feat-vada} and~\ref{fig:feat-vada-dirtt}),
and {\ours} models (Figures~\ref{fig:feat-gada} and ~\ref{fig:feat-gada-dirtt}).
We observe that the feature space of {\ours} is more organized with
more separate clusters, compared to those of VADA.
{\ours} increases the distance between clusters, which follows our intuition in the beginning.
This results in a much higher accuracy (83.6\% compared to 70.6\%).
When integrated with DIRT-T~\cite{shu2018a}, our performance becomes boosted even further from 83.6\% to 90\%.
This experiment shows the power of {\ours}, when integrated with other methods,
which proves the generic characteristic of our module.

\begin{table}[!t]
    \centering
    \caption{Accuracy on test set of the task MNIST $\rightarrow$ SVHN for ablation analysis.}
    \label{table:ablation}
    \begin{tabu}{ccccc|c}
        \tabucline[1pt]{-}
        $\Lc_c$    & $\Lc_d$    & $\Lc_e$    & $\Lc_v$    & $\Lc_u$    & MNIST $\rightarrow$ SVHN \\ \tabucline[1pt]{-}
        \checkmark & \checkmark &            &            &            & 66.3                     \\
        \checkmark & \checkmark & \checkmark &            &            & 68.1                     \\
        \checkmark & \checkmark &            & \checkmark &            & 69.9                     \\
        \checkmark & \checkmark &            &            & \checkmark & 78.7                     \\
        \checkmark & \checkmark & \checkmark & \checkmark &            & 70.6                     \\
        \checkmark & \checkmark & \checkmark & \checkmark & \checkmark & {\bf 83.6}               \\ \tabucline[1pt]{-}
    \end{tabu}
\end{table}

\myparagraph{Ablation study}
In order to understand the effects of each of loss functions in
our algorithm on the accuracy, we perform an extensive ablation
study by turning the losses on and off.
We test the loss functions on the challenging adaptation task
MNIST $\rightarrow$ SVHN.
Instance normalization is applied to all the cases in this analysis for fair comparison.
The ablation results are given in Table~\ref{table:ablation}.
The first row, where only $\Lc_c$ and $\Lc_d$ are used,
turns out to be  the result for our implementation of  DANN~\cite{ganin16}.
The next three lines show that adding one of $\Lc_e$, $\Lc_v$, or $\Lc_u$
into DANN improves the performance in a stable manner.
Among the three, $\Lc_u$ provides the highest improvement
($78.7\%$ compared to $68.1\%$ and $69.9\%$).
This improvement indicates that our module could be easily integrated
with other methods for higher performance.
We merge both $\Lc_e$ and $\Lc_v$ into DANN to have our implementation
of VADA~\cite{shu2018a}, which yields better performance than
when only one of them is integrated, as expected, though the performance gain
is still less than that of solely $\Lc_u$.
The best result is achieved when we add $\Lc_u$ into VADA,
which creates an improvement of $13\%$ in terms of accuracy
and surpasses the state-of-the-art result in CoDA~\cite{kumar18}.
This experiment, again, shows the power of our module
when integrated with other methods.

\begin{figure}[!t]
  \centering
  \includegraphics[width=\linewidth]{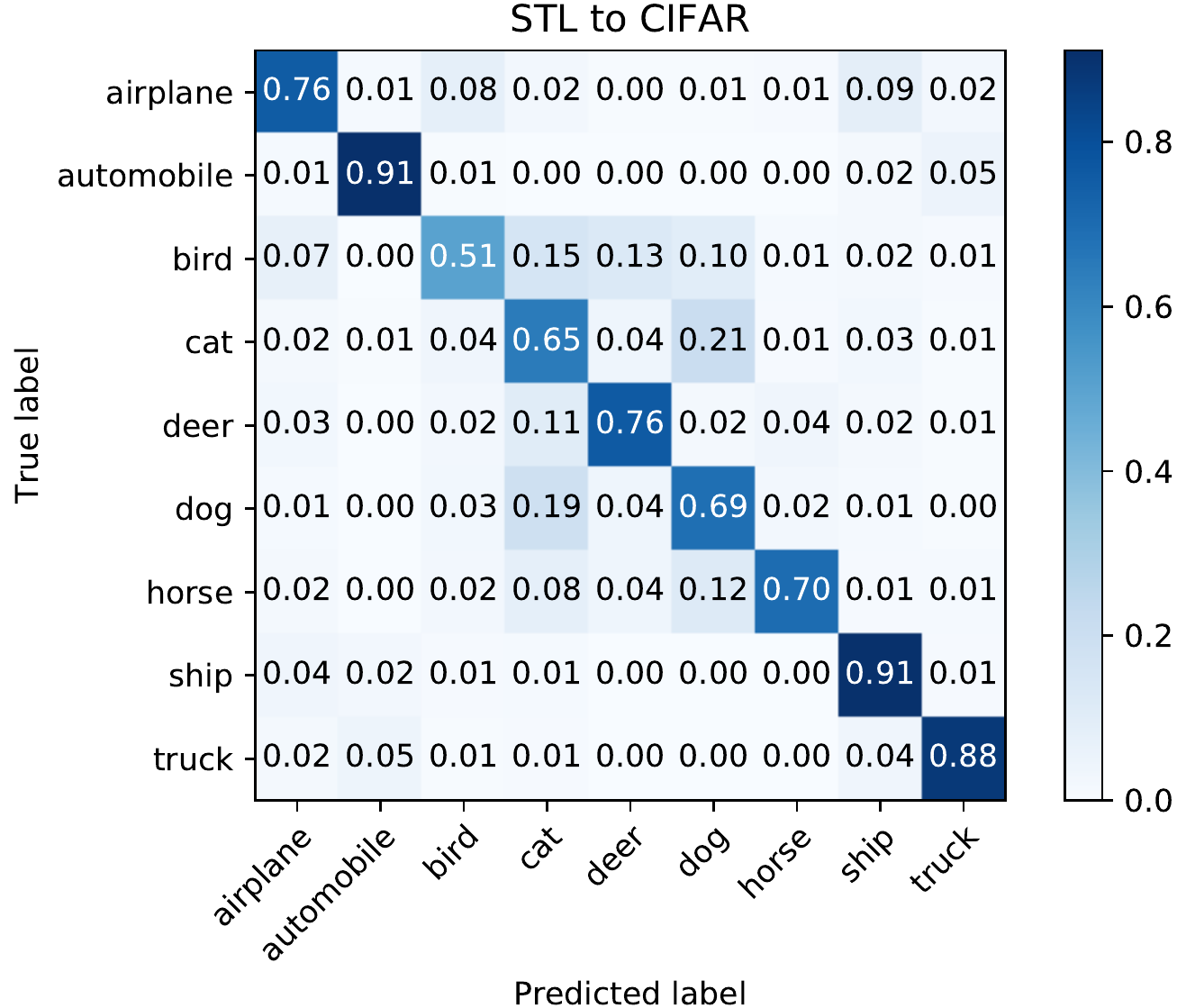}
  \caption{Confusion matrix for STL-10 $\rightarrow$ CIFAR-10.}
  \label{fig:confusion-matrix}
\end{figure}

\myparagraph{Confusion matrix}
In Figure~\ref{fig:confusion-matrix}, we present a confusion matrix
that shows the prediction accuracy for each of the nine different classes
in the task STL-10 $\rightarrow$ CIFAR-10.
We observe that our model works very well with several classes, such as `automobile,'
`ship,' and `truck,' each of them achieves accuracy of approximately 90\%.
The class that degrades our performance most is `bird' with only 51\% of accuracy.
Our model misclassifies the bird images as `cat', `deer', and `dog'.
We suspect that it is because of the noisy learning in the beginning of the training. The number of labels we have for the classification task is small,
which incorrectly moves samples to the wrong clusters.

\section{Conclusion}
\label{sec:conclusion}

We proposed the Generative Adversarial Domain Adaptation (GADA) algorithm, which significantly improves
the discriminative feature extraction process by injecting an extra class and training with generated samples.
The loss functions we proposed have the effects of separating the real target clusters,
therby helping the classifier easily find low-density areas to put the decision boundary into.
Through extensive experiments on different standard datasets, we showed the effectiveness of our method,
and outperformed the other state-of-the-art algorithms in many cases,
especially on the highly challenging adaptation task MNIST $\rightarrow$ SVHN.
In addition, our module is proved to be extremely effective when integrated into other methods.


\clearpage

\bibliographystyle{ieee_fullname}
\bibliography{ref}

\begin{thebibliography}{10}\itemsep=-1pt

\bibitem{amodei16}
Dario Amodei, Sundaram Ananthanarayanan, Rishita Anubhai, Jingliang Bai, Eric
  Battenberg, Carl Case, Jared Casper, Bryan Catanzaro, Qiang Cheng, Guoliang
  Chen, et~al.
\newblock Deep speech 2: End-to-end speech recognition in english and mandarin.
\newblock In {\em International conference on machine learning}, pages
  173--182, 2016.

\bibitem{bsds500}
Pablo Arbelaez, Michael Maire, Charless Fowlkes, and Jitendra Malik.
\newblock Contour detection and hierarchical image segmentation.
\newblock {\em IEEE transactions on pattern analysis and machine intelligence},
  33:898--916, 05 2011.

\bibitem{bou17}
Konstantinos Bousmalis, Nathan Silberman, David Dohan, Dumitru Erhan, and Dilip
  Krishnan.
\newblock Unsupervised pixel-level domain adaptation with generative
  adversarial networks.
\newblock In {\em 2017 IEEE Conference on Computer Vision and Pattern
  Recognition (CVPR)}, pages 95--104, 2017.

\bibitem{dsn16}
Konstantinos Bousmalis, George Trigeorgis, Nathan Silberman, Dilip Krishnan,
  and Dumitru Erhan.
\newblock Domain separation networks.
\newblock In {\em Advances in Neural Information Processing Systems}, pages
  343--351, 2016.

\bibitem{dai17}
Zihang Dai, Zhilin Yang, Fan Yang, William~W Cohen, and Ruslan~R Salakhutdinov.
\newblock Good semi-supervised learning that requires a bad gan.
\newblock In {\em Advances in Neural Information Processing Systems 30}, pages
  6510--6520, 2017.

\bibitem{ganin15}
Yaroslav Ganin and Victor Lempitsky.
\newblock Unsupervised domain adaptation by backpropagation.
\newblock In {\em Proceedings of the 32nd International Conference on Machine
  Learning}, pages 1180--1189, 2015.

\bibitem{ganin16}
Yaroslav Ganin, Evgeniya Ustinova, Hana Ajakan, Pascal Germain, Hugo
  Larochelle, Fran{\c{c}}ois Laviolette, Mario Marchand, and Victor Lempitsky.
\newblock Domain-adversarial training of neural networks.
\newblock {\em The Journal of Machine Learning Research}, 17(1):2096--2030,
  2016.

\bibitem{gan14}
Ian Goodfellow, Jean Pouget-Abadie, Mehdi Mirza, Bing Xu, David Warde-Farley,
  Sherjil Ozair, Aaron Courville, and Yoshua Bengio.
\newblock Generative adversarial nets.
\newblock In {\em Advances in neural information processing systems}, pages
  2672--2680, 2014.

\bibitem{he2016deep}
Kaiming He, Xiangyu Zhang, Shaoqing Ren, and Jian Sun.
\newblock Deep residual learning for image recognition.
\newblock In {\em Proceedings of the IEEE conference on computer vision and
  pattern recognition}, pages 770--778, 2016.

\bibitem{hoffman18a}
Judy Hoffman, Eric Tzeng, Taesung Park, Jun-Yan Zhu, Phillip Isola, Kate
  Saenko, Alexei Efros, and Trevor Darrell.
\newblock {C}y{CADA}: Cycle-consistent adversarial domain adaptation.
\newblock In {\em Proceedings of the 35th International Conference on Machine
  Learning}, pages 1989--1998, 2018.

\bibitem{kumar18}
Abhishek Kumar, Prasanna Sattigeri, Kahini Wadhawan, Leonid Karlinsky, Rogerio
  Feris, Bill Freeman, and Gregory Wornell.
\newblock Co-regularized alignment for unsupervised domain adaptation.
\newblock In {\em Advances in Neural Information Processing Systems}, pages
  9345--9356, 2018.

\bibitem{long15}
Mingsheng Long, Yue Cao, Jianmin Wang, and Michael~I. Jordan.
\newblock Learning transferable features with deep adaptation networks.
\newblock In {\em Proceedings of the 32Nd International Conference on
  International Conference on Machine Learning - Volume 37}, pages 97--105,
  2015.

\bibitem{long17}
Mingsheng Long, Han Zhu, Jianmin Wang, and Michael~I. Jordan.
\newblock Deep transfer learning with joint adaptation networks.
\newblock In {\em Proceedings of the 34th International Conference on Machine
  Learning - Volume 70}, pages 2208--2217, 2017.

\bibitem{murez18}
Zak Murez, Soheil Kolouri, David~J. Kriegman, Ravi Ramamoorthi, and Kyungnam
  Kim.
\newblock Image to image translation for domain adaptation.
\newblock In {\em 2018 IEEE Conference on Computer Vision and Pattern
  Recognition (CVPR)}, pages 4500--4509, 2018.

\bibitem{qi18}
Guo-Jun Qi, Liheng Zhang, Hao Hu, Marzieh Edraki, Jingdong Wang, and Xian-Sheng
  Hua.
\newblock Global versus localized generative adversarial nets.
\newblock In {\em The IEEE Conference on Computer Vision and Pattern
  Recognition (CVPR)}, 2018.

\bibitem{radford16}
Alec Radford, Luke Metz, and Soumith Chintala.
\newblock Unsupervised representation learning with deep convolutional
  generative adversarial networks.
\newblock In {\em International Conference on Learning Representations}, 2016.

\bibitem{artem18}
Artem Rozantsev, Mathieu Salzmann, and Pascal Fua.
\newblock Beyond sharing weights for deep domain adaptation.
\newblock {\em IEEE Transactions on Pattern Analysis and Machine Intelligence},
  41:801--814, 2018.

\bibitem{saito17a}
Kuniaki Saito, Yoshitaka Ushiku, and Tatsuya Harada.
\newblock Asymmetric tri-training for unsupervised domain adaptation.
\newblock In {\em Proceedings of the 34th International Conference on Machine
  Learning}, pages 2988--2997, 2017.

\bibitem{saito18}
Kuniaki Saito, Kohei Watanabe, Yoshitaka Ushiku, and Tatsuya Harada.
\newblock Maximum classifier discrepancy for unsupervised domain adaptation.
\newblock In {\em 2018 IEEE Conference on Computer Vision and Pattern
  Recognition (CVPR)}, pages 3723--3732, 2018.

\bibitem{salimans16}
Tim Salimans, Ian Goodfellow, Wojciech Zaremba, Vicki Cheung, Alec Radford, Xi
  Chen, and Xi Chen.
\newblock Improved techniques for training gans.
\newblock In {\em Advances in Neural Information Processing Systems 29}, pages
  2234--2242, 2016.

\bibitem{shu2018a}
Rui Shu, Hung Bui, Hirokazu Narui, and Stefano Ermon.
\newblock A {DIRT}-t approach to unsupervised domain adaptation.
\newblock In {\em International Conference on Learning Representations}, 2018.

\bibitem{sun16}
Baochen Sun and Kate Saenko.
\newblock Deep coral: Correlation alignment for deep domain adaptation.
\newblock In {\em Computer Vision -- ECCV 2016 Workshops}, pages 443--450,
  2016.

\bibitem{adda17}
Eric Tzeng, Judy Hoffman, Kate Saenko, and Trevor Darrell.
\newblock Adversarial discriminative domain adaptation.
\newblock In {\em {IEEE} Conference on Computer Vision and Pattern Recognition,
  {CVPR} 2017}, pages 2962--2971, 2017.

\bibitem{tzeng14}
Eric Tzeng, Judy Hoffman, Ning Zhang, Kate Saenko, and Trevor Darrell.
\newblock Deep domain confusion: Maximizing for domain invariance.
\newblock {\em CoRR}, abs/1412.3474, 2014.

\bibitem{gagl18}
Kai-Ya Wei and Chiou-Ting Hsu.
\newblock Generative adversarial guided learning for domain adaptation.
\newblock In {\em British Machine Vision Conference (BMVC)}, 2018.

\bibitem{xie18c}
Shaoan Xie, Zibin Zheng, Liang Chen, and Chuan Chen.
\newblock Learning semantic representations for unsupervised domain adaptation.
\newblock In {\em Proceedings of the 35th International Conference on Machine
  Learning}, pages 5423--5432, 2018.

\bibitem{cmd17}
Werner Zellinger, Thomas Grubinger, Edwin Lughofer, Thomas Natschläger, and
  Susanne Saminger-Platz.
\newblock Central moment discrepancy (cmd) for domain-invariant representation
  learning.
\newblock In {\em International Conference on Learning Representations}, 2017.

\bibitem{cyclegan}
Jun{-}Yan Zhu, Taesung Park, Phillip Isola, and Alexei~A. Efros.
\newblock Unpaired image-to-image translation using cycle-consistent
  adversarial networks.
\newblock In {\em {IEEE} International Conference on Computer Vision, {ICCV}
  2017, Venice, Italy, October 22-29, 2017}, pages 2242--2251, 2017.

\end{thebibliography}

\onecolumn

\newcommand{\horrule}[1]{\rule{\linewidth}{#1}}

{
\center
\horrule{0.5pt} \\[0.4cm]
\huge Supplementary Materials \\
\horrule{2pt} \\[0.5cm]
}

\newcommand{\beginsupplement}{%
    \setcounter{section}{0}
    \renewcommand{\thesection}{S\arabic{section}}%
    \setcounter{table}{0}
    \renewcommand{\thetable}{S\arabic{table}}%
    \setcounter{figure}{0}
    \renewcommand{\thefigure}{S\arabic{figure}}%
}

\beginsupplement

\section{Network architectures}
In Table~\ref{table:main-classifier}, we present the network
architectures of the main classifier, which has a small version for digit
datasets (MNIST, SVHN, MNIST-M, SynthDigits) and a large one
for object datasets (CIFAR-10 and STL-10).
Domain discriminator architecture is in Table~\ref{table:domain-discriminator},
which is the same for both small and large classifiers.
Generator architecture used in the end-to-end training scheme for digit datasets
is presented in Table~\ref{table:gen-small}.
Recall that for the object datasets, we perform a pretraining stage to train the generator.
This stage uses a generator architecture in Table~\ref{table:gen-large} and a discriminator
architecture in Table~\ref{table:discriminator-gan}.
The architecture and training scheme in this pretraining stage partially follows DCGAN~\cite{radford16}.

Please note in the task SVHN $\rightarrow$ MNIST, the input features to the domain discriminator $D$
is chosen to be the last layer of the classifier $f$ (Layer 15, before softmax),
and the feature layer chosen to calculate the mean for generator training is Layer 13.
For all other tasks, input features to the domain discriminator $D$ is from Layer 13,
and the feature mean for generator training is calculated using output from Layer 14.

\begin{table}[!ht]
    \centering
    \small
    \caption{Network architecture for the main classifier $f$.
    `SAME' and `VALID' indicate the padding scheme used in each convolutional layer.
    Batch normalization is applied before activation of all convolutional and dense layers.
    Leaky ReLU parameter $\alpha$ is set to 0.1.
    The use of the additive Gaussian noise is empirically proved to improve the performance.}
    \label{table:main-classifier}
    \begin{tabu}{c|cc}
        \tabucline[1pt]{-}
        Layer Index & SMALL NETWORK                                 & LARGE NETWORK                                 \\ \tabucline[1pt]{-}
        0           & \multicolumn{2}{c}{$32\times 32 \times 3$ input images}                                       \\ \tabucline[.5pt]{-}
        1           & \multicolumn{2}{c}{Instance Normalization (optional)}                                         \\ \tabucline[.5pt]{-}
        2           & $32\times 3 \times 3$ Conv (SAME), lReLU      & $96\times 3 \times 3$ Conv (SAME), lReLU      \\
        3           & $32\times 3 \times 3$ Conv (SAME), lReLU      & $96\times 3 \times 3$ Conv (SAME), lReLU      \\
        4           & $32\times 5 \times 5$ Conv (VALID), lReLU     & $96\times 5 \times 5$ Conv (VALID), lReLU     \\ \tabucline[.5pt]{-}
        5           & \multicolumn{2}{c}{$2\times 2$ max-pooling, stride 2}                                         \\
        6           & \multicolumn{2}{c}{Dropout, $p=0.5$}                                                          \\
        7           & \multicolumn{2}{c}{Gaussian noise, $\sigma=1$}                                                \\ \tabucline[.5pt]{-}
        8           & $64\times 3 \times 3$ Conv (SAME), lReLU      & $192\times 5 \times 5$ Conv (SAME), lReLU     \\
        9           & $64\times 3 \times 3$ Conv (SAME), lReLU      & $192\times 5 \times 5$ Conv (SAME), lReLU     \\
        10          & $64\times 5 \times 5$ Conv (VALID), lReLU     & $192\times 5 \times 5$ Conv (VALID), lReLU    \\ \tabucline[.5pt]{-}
        11          & \multicolumn{2}{c}{$2\times 2$ max-pooling, stride 2}                                         \\
        12          & \multicolumn{2}{c}{Dropout, $p=0.5$}                                                          \\
        13          & \multicolumn{2}{c}{Gaussian noise, $\sigma=1$}                                                \\ \tabucline[.5pt]{-}
        14          & Dense 500, lReLU                              & Dense 2048, lReLU                             \\ \tabucline[.5pt]{-}
        15          & \multicolumn{2}{c}{Dense output, softmax}                                                     \\ \tabucline[1pt]{-}
    \end{tabu}
\end{table}

\begin{table}[!ht]
    \centering
    \small
    \caption{Network architecture for the domain discriminator $D$.
    Input features are the output of Layer 15 before softmax (in SVHN $\rightarrow$ MNIST)
    or the output of Layer 13 (in all other tasks) from the main classifier $f$.}
    \label{table:domain-discriminator}
    \begin{tabu}{c|c}
        \tabucline[1pt]{-}
        Layer Index & Domain Discriminator  \\ \tabucline[1pt]{-}
        0           & Input features        \\
        1           & Dense 500, ReLU       \\
        2           & Dense 100, ReLU       \\
        3           & Dense 1, sigmoid      \\ \tabucline[1pt]{-}
    \end{tabu}
\end{table}

\begin{table}[!ht]
    \centering
    \small
    \caption{Generator architecture used for the small classifier.
    All transposed convolutional layers use the `SAME' padding scheme.
    Batch normalization is applied before activation of all transposed convolutional
    and dense layers, except for the output layer.
    Leaky ReLU parameter $\alpha$ is set to 0.1.}
    \label{table:gen-small}
    \begin{tabu}{c|c}
        \tabucline[1pt]{-}
        Layer Index & Discriminator                                             \\ \tabucline[1pt]{-}
        0           & Input noise vector of length $100$                        \\
        1           & Dense 8192, reshape to $512\times 4\times 4$, lRELU       \\
        2           & $256\times 3 \times 3$ Transposed Conv, stride 2, lRELU   \\
        3           & $128\times 3 \times 3$ Transposed Conv, stride 2, lRELU   \\
        4           & $3 \times 3 \times 3$ Transposed Conv, stride 2, tanh     \\ \tabucline[1pt]{-}
    \end{tabu}
\end{table}

\begin{table}[!ht]
    \centering
    \small
    \caption{Generator architecture used in pretraining and then for the large classifier.
    All transposed convolutional layers use the `SAME' padding scheme.
    Batch normalization is applied before activation of all transposed convolutional
    and dense layers, except for the output layer.}
    \label{table:gen-large}
    \begin{tabu}{c|c}
        \tabucline[1pt]{-}
        Layer Index & Discriminator                                             \\ \tabucline[1pt]{-}
        0           & Input noise vector of length $100$                        \\
        1           & Dense 2048, reshape to $512\times 2\times 2$, ReLU        \\
        2           & $256\times 5 \times 5$ Transposed Conv, stride 2, RELU    \\
        3           & $128\times 5 \times 5$ Transposed Conv, stride 2, RELU    \\
        4           & $64\times 5 \times 5$ Transposed Conv, stride 2, RELU     \\
        5           & $3 \times 5 \times 5$ Transposed Conv, stride 2, tanh     \\ \tabucline[1pt]{-}
    \end{tabu}
\end{table}

\begin{table}[!ht]
    \centering
    \small
    \caption{Network architecture for the discriminator used in GAN pretraining.
    All convolutional layers use the `SAME' padding scheme.
    Batch normalization is applied before activation of all transposed convolutional
    and dense layers, except for the first convolutional layer and the output layer.
    Leaky ReLU parameter $\alpha$ is set to 0.1.}
    \label{table:discriminator-gan}
    \begin{tabu}{c|c}
        \tabucline[1pt]{-}
        Layer Index & Discriminator                                 \\ \tabucline[1pt]{-}
        0           & $32\times 32 \times 3$ input images           \\
        1           & $64\times 5\times 5$ Conv, stride 2,  lRELU   \\
        2           & $128\times 5\times 5$ Conv, stride 2, lRELU   \\
        3           & $256\times 5\times 5$ Conv, stride 2, lRELU   \\
        4           & $512\times 5\times 5$ Conv, stride 2, lRELU   \\
        5           & Dense 1, sigmoid                              \\ \tabucline[1pt]{-}
    \end{tabu}
\end{table}

\section{Hyperparameters}

In all experiments, we restrict the hyperparameter search to
$\lambda_d=\{10^{-2}, 0\}$,
$\lambda_s=\{0,1\}$,
$\lambda_t=\{10^{-1},10^{-2}\}$,
$\lambda_u=\{10^{-1},1\}$,
and $\text{lr}=\{2\times10^{-4},10^{-3}\}$.
The set of hyperparameters which work the best for each task
is given in the Table~\ref{table:hyperparams}.
We observe that $\lambda_u=1$ works constantly well on all the cases, and
changing this value to $10^{-1}$ does not affect the performance much.
Note that $\Lc_e$ and $\Lc_v$ are not used in the tasks
SVHN $\rightarrow$ MNIST and CIFAR-10 $\rightarrow$ STL-10
$\left(\lambda_s=\lambda_t=0\right)$.
$\lambda_d$ is set to $0$ in the tasks CIFAR-10 $\leftrightarrow$ STL-10
following the prior belief in~\cite{shu2018a}.
$\lambda_s$ is set to $0$ in the task STL-10 $\rightarrow$ CIFAR-10
because the number of source samples in this case (STL-10 images)
is very small, which is unreliable to learn from.
All trainings use Adam Optimizer with $\beta_1=0.5$ and $\beta_2=0.999$.

\begin{table}[!ht]
    \centering
    \caption{Hyperparameters used for each task.}
    \label{table:hyperparams}
    \begin{tabu}{c|ccccc}
        \tabucline[1pt]{-}
        Task                           & $\lambda_d$ & $\lambda_s$ & $\lambda_t$ & $\lambda_u$ & lr                 \\ \tabucline[1pt]{-}
        MNIST $\rightarrow$ SVHN       & $10^{-2}$   & 1           & $10^{-2}$   & 1           & $2\times 10^{-4}$  \\
        SVHN $\rightarrow$ MNIST       & $10^{-2}$   & 0           & 0           & 1           & $2\times 10^{-4}$  \\
        MNIST $\rightarrow$ MNIST-M    & $10^{-2}$   & 0           & $10^{-2}$   & 1           & $2\times 10^{-4}$  \\
        SynthDigits $\rightarrow$ SVHN & $10^{-2}$   & 1           & $10^{-2}$   & 1           & $2\times 10^{-4}$  \\
        CIFAR-10 $\rightarrow$ STL-10  & 0           & 0           & 0           & 1           & $10^{-3}$          \\
        STL-10 $\rightarrow$ CIFAR-10  & 0           & 0           & $10^{-1}$   & 1           & $10^{-3}$          \\ \tabucline[1pt]{-}
    \end{tabu}
\end{table}

Note that in both tasks CIFAR-10 $\rightarrow$ STL-10 and STL-10 $\rightarrow$ CIFAR-10,
we pretrain the generator before actually doing the main training phase.
During the pretraining, we train the generator with a learning rate of $10^{-4}$
and train the discriminator with a learning rate of $10^{-3}$.
After that, we use the pretrained generator to train the classifier $f$ for the domain adaptation tasks.
During this main training phase, we keep fine-tuning the generator with a small learning rate of $2\times 10^{-5}$.
In addition, for the task CIFAR-10 $\rightarrow$ STL-10, we pretrain the generator with only 15 epochs,
while for the task STL-10 $\rightarrow$ CIFAR-10, we use 100 epochs.
This is due to a large difference between number of samples in STL-10 dataset ($450$ images)
compared to that of CIFAR-10 ($45,000$ images).

\section{More on how OOC samples help separate the real clusters}

\subsection{Limitations of VADA and DIRT-T}
\begin{figure}[!ht]
  \centering
  \begin{subfigure}{0.5\textwidth}
    \centering
    \includegraphics[width=0.5\linewidth]{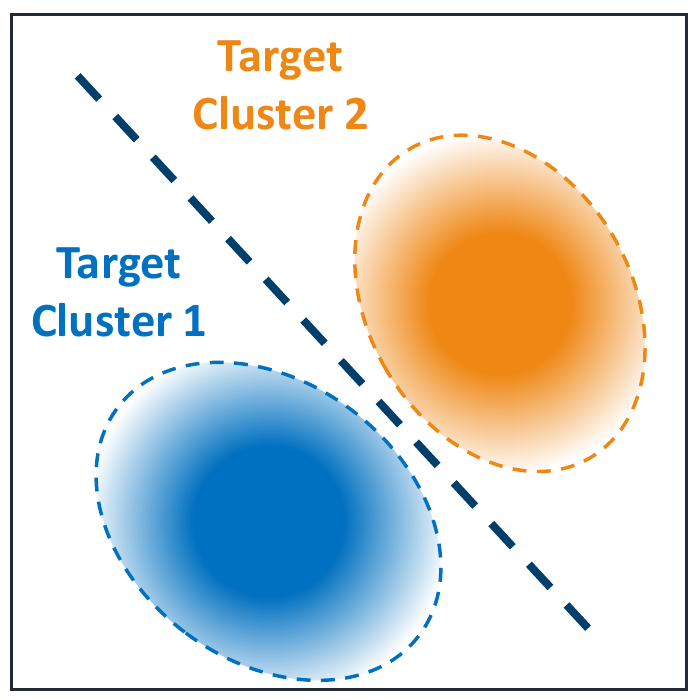}
    \caption{Perfect Scenario}
    \label{fig:perfect-case}
  \end{subfigure}%
  \hspace{-1.2cm}
  \begin{subfigure}{0.5\textwidth}
    \centering
    \includegraphics[width=0.5\linewidth]{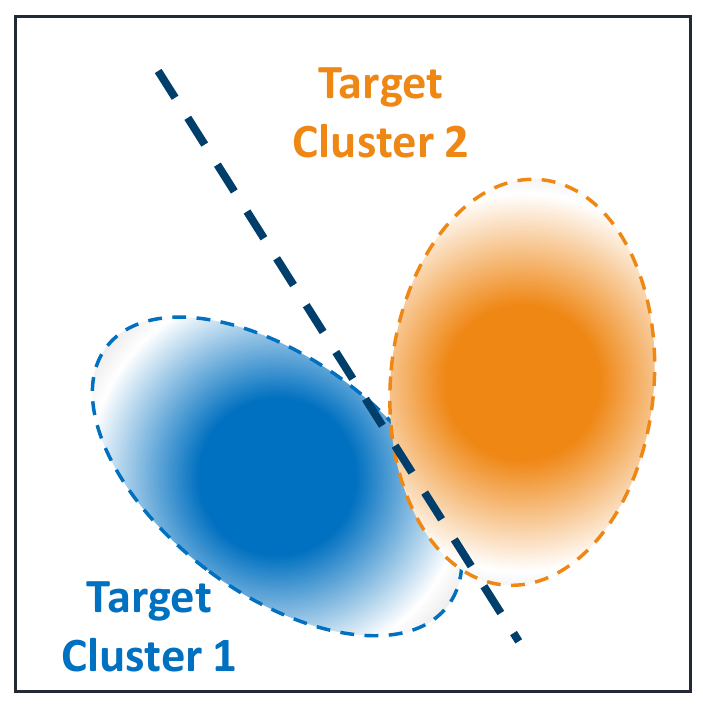}
    \caption{Real Scenario}
    \label{fig:real-case}
  \end{subfigure}
  \caption[Problem of VADA and DIRT-T]
  {Problem of VADA and DIRT-T~\cite{shu2018a}.
  The authors assume a perfect scenario as in
  Figure~\protect\subref{fig:perfect-case},
  while the scenario that will probably happen
  is in Figure~\protect\subref{fig:real-case}.}
\end{figure}

Recall that in VADA and DIRT-T methods~\cite{shu2018a}, the authors
tried to move the decision boundaries to low-density areas,
with the assumption that the features in the feature space
are located into separate clusters, where samples in the
same cluster share the same label.
However, a perfect scenario in Figure~\ref{fig:perfect-case},
where the target clusters are clearly separate so that the
decision boundary could go through the low-density area
between them, is not guaranteed because the model has
no information on the target labels.
This perfect scenario could happen with source
clusters, where we have excessive information about
labels, but not with target clusters.
A more realistic scenario is illustrated in Figure~\ref{fig:real-case},
where the target clusters have a small overlapping region, thereby
degrading the performance despite the fact that the decision boundary
is passing the low-density area between the two clusters.
In our research, the goal is to create larger areas between the target
clusters in the feature space so that they are clearly separated.
In other words, we want to increase the distance between the target
clusters in the feature space.

\subsection{How OOC samples help}

\begin{figure}[!ht]
  \centering
  \includegraphics[width=\linewidth]{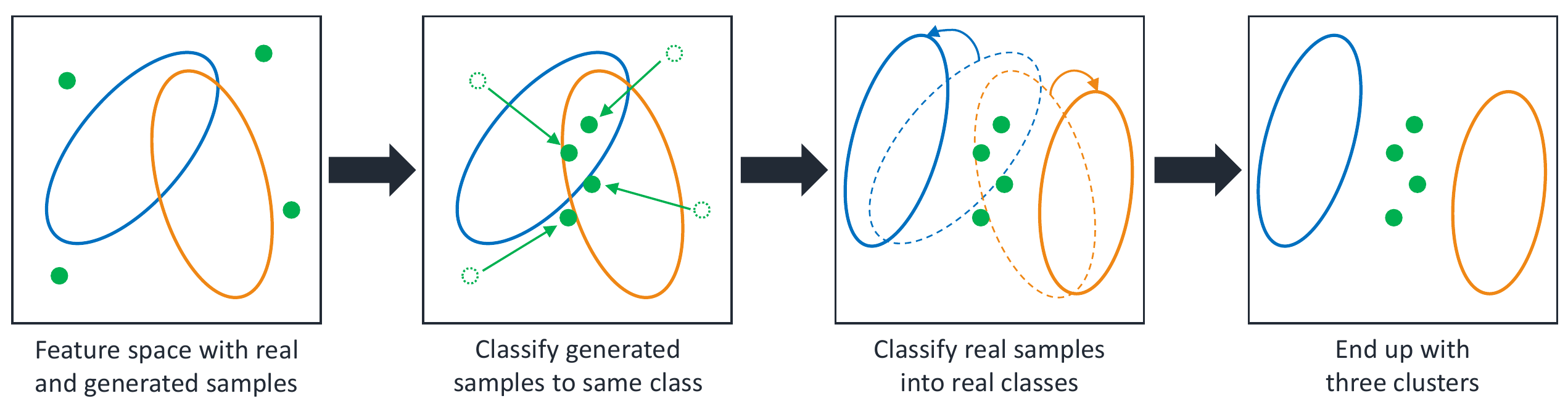}
  \caption{Intuition of how OOC samples help in the learning process.
  Suppose in the feature space extracted by a neural network,
  there are two overlapping feature clusters of real data and several
  green feature points of generated samples ($1^\text{st}$~figure).
  All the generated samples should belong to the same fake $\text{(K+1)}^\text{th}$~class,
  so they will be moved closer to each other into a same new cluster ($2^\text{nd}$~figure).
  The two real-class-feature clusters belong to the real classes, so they
  should not overlap with the features of the generated samples.
  Therefore, they will be moved away from the position of
  generated-sample features in the middle ($3^\text{rd}$~figure).
  After all mentioned learning steps, the feature space ends up
  with the fake cluster being placed in the middle of the two real clusters ($4^\text{th}$~figure).
  }
  \label{fig:ooc-samples-intuition}
\end{figure}

In our model, the classifier $f$ must distinguish between the real
and fake samples in order to make the real clusters more separated,
thereby improving the performance.
An illustration is given in Figure~\ref{fig:ooc-samples-intuition} to explain this statement.
In this figure, we see that the feature space ends up with
the fake cluster being placed in the middle of the two real clusters.
This scenario increases the distance between the real clusters,
thereby achieving the ultimate goal.

One might argue that the fake cluster could be placed far
away from the real clusters.
However, this situation should not happen given that the
generated samples have features similar to the ones of real data.
For example, consider a generator trying to
generate numbers from the MNIST dataset.
Suppose in the feature space, real clusters of number
$1$ and of number $7$ are placed near each
other because of their similar shape.
Weird-shape numbers $1$ and $7$ are generated,
i.e., they come from a distribution different from the distribution
of normal-shape numbers $1$ and $7$ in the training data.
These generated numbers are ``bad'', but they are still $1$ and $7$,
which should be placed near the corresponding clusters in the feature space.
On the other hand, these generated samples belong to the same fake cluster.
Therefore, they will be moved to a same cluster
while staying near the corresponding real clusters,
which results in a feature space similar to
the $4^\text{th}$ figure in Figure~\ref{fig:ooc-samples-intuition}.

\section{Generated images}

\begin{figure}[!ht]
  \centering
  \begin{subfigure}{0.5\textwidth}
    \centering
    \includegraphics[width=0.8\linewidth]{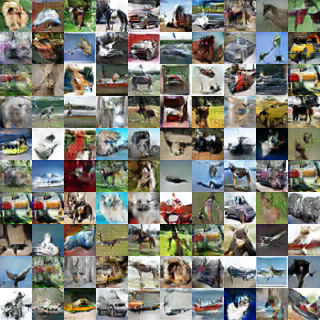}
    \caption{Generated CIFAR-10 images}
    \label{fig:cifar-gen}
  \end{subfigure}%
  \hspace{-1.2cm}
  \begin{subfigure}{0.5\textwidth}
    \centering
    \includegraphics[width=0.8\linewidth]{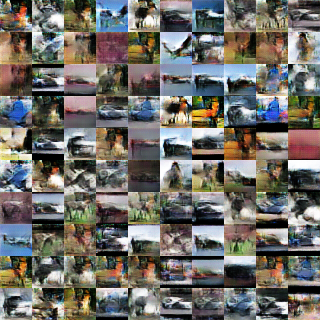}
    \caption{Generated STL-10 images}
    \label{fig:stl-gen}
  \end{subfigure}
  \caption{Generated images on the task
  \subref{fig:cifar-gen} STL-10 $\rightarrow$ CIFAR-10
  and \subref{fig:stl-gen} CIFAR-10 $\rightarrow$ STL-10.}
  \label{fig:gen-imgs}
\end{figure}

We show some of the generated images on the tasks
CIFAR-10 $\rightarrow$ STL-10 and STL-10 $\rightarrow$ CIFAR-10.
We could observe that the generated samples on CIFAR-10
in Figure~\ref{fig:cifar-gen} have higher quality in terms
of variety, compared to STL-10 generated images in Figure~\ref{fig:stl-gen}.
This is because of a large difference between the numbers of unsupervised
training samples given in each task.
Number of STL-10 target samples is much lower than that of the CIFAR-10 dataset:
$450$ compared to $45,000$ respectively.

\end{document}